\documentclass{article}

\usepackage[main, final]{main}

\usepackage[utf8]{inputenc} 
\usepackage[T1]{fontenc}    
\usepackage[colorlinks,
            linkcolor=black,       
            citecolor=blue!50!black,     
            urlcolor=blue        
            ]{hyperref}
\usepackage{url}            
\usepackage{booktabs}       
\usepackage{amsfonts}       
\usepackage{amssymb}
\usepackage{authblk}
\usepackage{nicefrac}       
\usepackage{microtype}      
\usepackage[table]{xcolor}         
\usepackage{graphicx}
\usepackage{svg} 
\usepackage{float}
\usepackage{tikz}
\usepackage{geometry}
\usepackage{enumitem}
\labellogo{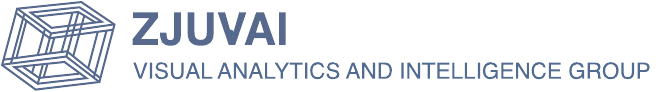}
\usetikzlibrary{positioning, fit, backgrounds}
\usepackage{cleveref}
\usepackage{sourcesanspro}
\usepackage[mode=buildnew]{standalone}
\usepackage{tikz}
\usetikzlibrary{shapes.geometric, arrows.meta, positioning, fit, calc, backgrounds, shadows}

\usepackage[commandnameprefix=always]{changes}
\newcommand{\mtil}[1]{%
    \raisebox{0pt}[0pt][0pt]{%
        \fboxsep=1pt 
        \colorbox{lightgray}{\texttt{#1}}%
    }%
}

\title{GenesisGeo: Technical Report}

\author[1]{Minfeng Zhu}
\author[1]{Zi Wang}
\author[1]{Sizhe Ji}
\author[1]{Zhengtong Du}
\author[2]{Shengqiang Tai}
\author[3]{Junming Ke}
\author[3]{Xiao Deng}
\author[4]{Zanlang Yin}
\author[1]{Xiuqi Huang}
\author[5]{Heyu Wang}
\author[1]{Wei Chen}

\affil[1]{State Key Lab of CAD\&CG, Zhejiang University}
\affil[2]{Polytechnic Institute, Zhejiang University}
\affil[3]{Hangzhou Research Institute of AI and Holographic Technology}
\affil[4]{Volkswagen Group Innovation}
\affil[5]{School of Mathematical Science, Zhejiang University}

\affil[ ]{\fontfamily{lmtt}\selectfont \{minfeng\_zhu, zi\_wang, sizhe\_ji, zhengtongdu, shengqiangtai, huangxiuqi, wangheyu, chenvis\}@zju.edu.cn}
\affil[ ]{\fontfamily{lmtt}\selectfont \{junmingke, xiaodeng\}@zjuqx.com, \fontfamily{lmtt}\selectfont zanlang.yin@gmail.com}

\begin{document}
\maketitle

\begin{abstract}
Recent neuro-symbolic geometry theorem provers have made significant progress on Euclidean problems by coupling neural guidance with symbolic verification. However, most existing systems operate almost exclusively in a symbolic space, leaving diagram-based intuition largely unused during reasoning. For humans, geometric diagrams provide essential heuristics for identifying non-trivial auxiliary constructions. Meanwhile, visual language models (VLMs) still struggle with geometry due to the lack of high-quality data with geometric diagrams and reasoning supervision.
In this paper, we introduce \textsc{GenesisGeo-1M}, a large-scale synthetic dataset for visual geometric reasoning that contains 1M multimodal geometry problems paired with machine-checkable proof traces. Building on this dataset, we formulate geometric learning as a multi-task training paradigm that jointly optimizes text-based proof generation and diagram-grounded proof generation, encouraging models to learn visual grounding and symbolic deduction. Extensive experiments show that our \textsc{GenesisGeo-2B} model achieves gold-medal-level performance on Olympiad geometry benchmarks, solving 29/30 problems on IMO-30, 63/95 on IMO-95, and 278/409 on HAGeo-409.

\end{abstract}

\section{Introduction}
Automated theorem proving in Euclidean geometry has made remarkable progress in recent years~\cite{trinh2024alphageometry,duan2025hagio,chen2025seedproverdeepbroadreasoning}. Neuro-symbolic systems such as AlphaGeometry2 can reach gold-medal-level performance~\cite{chervonyi2025alphageometry2}. These systems show that combining language-guided auxiliary construction search with symbolic deduction can support long-horizon geometric proofs.
However, most existing methods solve problems primarily in a textual space. They encode the problem in a formal language, guide auxiliary construction search using language- or heuristic-based priors, and verify conclusions via symbolic deduction. This workflow overlooks visual cues (e.g., symmetry, alignment, concurrency, and cyclic patterns) that often suggest the correct auxiliary construction in Olympiad geometry.

In human problem-solving, diagrams serve as a perceptual workspace for forming hypotheses, checking constraints, and identifying useful auxiliary constructions. Experienced mathematicians often use visual cues to choose auxiliary constructions before writing a proof. This visual intuition is central to Olympiad geometry, where a solution often depends on adding the right auxiliary point, line, or circle.
In contrast, current VLMs remain unreliable on geometry problems with diagrams~\cite{10.1145/3746027.3754571,lu2023mathvista}. One reason is the lack of large-scale multimodal geometry data that pairs diagrams with machine-checkable proof traces. Existing multimodal datasets rarely expose models to the full reasoning pipeline required for Olympiad geometry, from perceptual pattern recognition to auxiliary construction and formal verification. This disconnect prevents models from learning how visual evidence informs symbolic planning in long-horizon proofs. As a result, current systems often fail to discover non-obvious auxiliary constructions that are critical for solving competition-level problems.

To address this gap, we introduce \textsc{GenesisGeo-1M}, a large-scale synthetic dataset for visual geometric reasoning with verified proof supervision. We enable this scale by reimplementing the symbolic deduction engine in C++ and equipping it with a carefully curated rule set (e.g., the \emph{secant theorem}) that improves both deductive capability and runtime efficiency.
Building on this engine, we develop an automated synthesis pipeline that generates challenging Olympiad-style problems. The pipeline samples construction-based problems, performs symbolic reasoning to obtain a formally checkable proof trace with the necessary auxiliary constructions, and renders the geometry constructions into a diagram with deterministic grounding of symbolic objects in the visual space. Each example includes a problem statement, a diagram, and a step-by-step symbolic proof trace. This design tightly aligns visual representations, symbolic premises, auxiliary constructions, and proof steps, enabling models to learn how perceptual cues inform formal deduction.
Compared to prior datasets~\cite{gao2025gllava,wu2025nesygeoneurosymbolicframeworkmultimodal}, \textsc{GenesisGeo-1M} provides large-scale multimodal geometry problems that enable long-horizon, diagram-grounded geometric reasoning.

In addition, we introduce \textsc{GenesisGeo-2B}, a multimodal geometry model trained with a multi-task learning strategy, including text-based proof generation, and diagram-grounded proof generation. 
These tasks share the same underlying geometry but provide complementary supervision signals: text-based proof generation strengthens symbolic rigor without relying on visual shortcuts, and diagram-grounded proof generation teaches the model to use visual evidence when proposing auxiliary constructions.
We fine-tune strong vision-language backbones on \textsc{GenesisGeo-1M} under this multi-task training strategy. Experiments show consistent gains on established geometry benchmarks. In particular, \textsc{GenesisGeo-2B} solves 29/30 problems on IMO-30, 63/95 on IMO-95, and 278/409 on HAGeo-409, suggesting that large-scale multimodal supervision improves diagram understanding and symbolic reasoning.

In summary, our contributions are fourfold:
\begin{itemize}[itemsep=2pt,topsep=0pt,parsep=0pt,leftmargin=10pt]
    \item We release \textsc{GenesisGeo-1M}, a multimodal geometry dataset with diagrams and machine-checkable proof traces for visual geometric reasoning.
    \item We propose a multi-task training strategy that jointly learns diagram understanding, text-based proof generation, and diagram-grounded proof generation.
    \item We conduct extensive experiments on established geometry benchmarks and demonstrate substantial gains after fine-tuning on \textsc{GenesisGeo-1M}.
    \item We provide a new solution to IMO 2008 Problem~6 that requires only two auxiliary points, simplifying AlphaGeometry-based solutions.
\end{itemize}
\section{Related Work}

\subsection{Geometry Problem Solving}
Automated geometric reasoning has a long history, progressing from algebraic elimination to symbolic proof systems and, more recently, neuro-symbolic approaches.

\textbf{Algebraic and symbolic solvers.}
Early works on automated geometry translate geometric constraints into algebraic elimination, including Wu’s method and related mechanical theorem-proving systems~\cite{wu1978method,chou1988mechanical}, as well as Gröbner-basis procedures~\cite{KAPUR1986399}. These algebraic approaches can be computationally efficient, but the resulting proofs are often difficult for humans to interpret. Alternative lines of work aimed to produce human-readable proofs, notably the area method~\cite{chou1993area} and deductive databases~\cite{chou2000deductive}. Recent work shows that hand-crafted symbolic heuristics can outperform classic methods on Olympiad-level benchmarks~\cite{duan2025hagio}.
However, purely symbolic search can become inefficient when auxiliary constructions rapidly expand the search space.

\textbf{Neuro-symbolic solvers.}
Recent approaches address search explosion by integrating deep neural networks for auxiliary construction proposal. AlphaGeometry couples a deductive database with an algebraic engine and trains a language model on synthetic textual corpora to suggest auxiliary points~\cite{trinh2024alphageometry}. Subsequent systems enhance the symbolic engine and data construction (e.g., TongGeometry~\cite{zhang2024tonggeometry}, SeedProver~\cite{chen2025seedproverdeepbroadreasoning}, and AlphaGeometry2~\cite{chervonyi2025alphageometry2}), solving the full IMO-30 set and achieving high success rates over broader collections of IMO problems. For example, TongGeometry applies tree search to explore auxiliary constructions~\cite{zhang2024tonggeometry}. AlphaGeometry2 extends the domain-specific language with dynamic point moves and linear relations, achieving an 84\% success rate on IMO problems from 2000--2024~\cite{chervonyi2025alphageometry2}. Despite these advances, most approaches still operate on symbolic inputs, without exploiting rich visual cues. We aim to learn construction-relevant auxiliary proposals from visual diagram evidence.

\subsection{Multimodal Geometric Reasoning}
Multimodal models such as ChatGPT and Gemini perform well on general vision tasks but remain unreliable at parsing geometric diagrams and performing multi-step reasoning~\cite{wang2024measuring}. Specialized models aim to close this gap by improving either \emph{understanding} or \emph{reasoning}.

\textbf{Understanding-oriented} work improves the mapping from pixels to geometric primitives and relations.
G-LLaVA~\cite{gao2025gllava} synthesizes geometric vision-language data to provide MLLMs with fundamental geometric knowledge.
GeoVLMath uses cross-modal rewards to better ground auxiliary-line descriptions in the diagram~\cite{guo2025geovlmath}. MathMathCoder-VL, trained with ImgCode-8.6M, leverages code as supervision for cross-modal alignment and understanding~\cite{wang-etal-2025-mathcoder}.

\textbf{Reasoning-oriented} work introduces step-by-step supervision to encourage multi-step deduction. Datasets such as GeoThought and GPSM4K provide intermediate reasoning traces and improve performance under chain-of-thought style training~\cite{shi2025geothought,anand2024gpsm4k}. GeoGen highlights hallucination issues in direct multimodal reasoning and proposes neuro-symbolic integration with translation and verification components~\cite{10.1145/3746027.3754571}.

In contrast to prior work, we aim to push multimodal geometric reasoning to IMO-level difficulty by leveraging a multi-task training strategy that learns geometric visual cues while a symbolic engine verifies logical soundness.

\subsection{Geometry Datasets}
Geometry datasets broadly fall into (1) textural datasets that teach symbolic deduction and (2) multimodal datasets that pair diagrams with questions and proof traces.

\textbf{Textural datasets.}
Large synthetic corpora have been central to recent neuro-symbolic provers. AlphaGeometry trains language models on 100M problems to learn auxiliary-construction priors and guide symbolic search~\cite{trinh2024alphageometry}. Follow-up systems scale the generation pipeline and search procedure (e.g., TongGeometry~\cite{zhang2024tonggeometry} and GeoGen~\cite{bak2020automated}) to further improve coverage and robustness. These corpora are effective for geometric reasoning in symbolic space, but they omit diagrams and therefore cannot teach the perceptual cues that humans rely on when selecting constructions.

\textbf{Multimodal datasets.}
Human-curated benchmarks such as Geometry3K and GeoQA provide diagrams paired with formal annotations or executable programs~\cite{lu2021intergps,chen2021geoqa}. Further, UniGeo unifies calculation and proving tasks with proving and program sequences that shares the same format~\cite{chen-etal-2022-unigeo}. AuxSolidMath curates exam-style problems with paired diagrams and aligned text for auxiliary-line reasoning~\cite{guo2025geovlmath}. 
For scaling multimodal training, Geo170K supplies diagram-text pairs~\cite{gao2025gllava} and GeoThought further provide detailed thinking processes during problem-solving~\cite{shi2025geothought}. Recently, researchers employ a neuro-symbolic framework for generating geometric reasoning data (e.g., NeSyGeo~\cite{wu2025nesygeoneurosymbolicframeworkmultimodal}, GeoGen~\cite{10.1145/3746027.3754571}, TrustGeoGen~\cite{fu2025trustgeogenformalverifieddataengine}).

Despite rapid progress, the field still lacks large-scale multimodal training data at IMO-level difficulty. We introduce \textsc{GenesisGeo-1M} dataset with diagram-text pairs and machine-checkable proof traces, which enables gold-medal-level performance on Olympiad geometry benchmarks.

\section{Preliminary of Symbolic Engine}
This section introduces our symbolic engine, which operates over two geometry-specific DSLs, and performs forward closure via a Deductive Database and Algebraic Reasoning (DDAR) procedure. We also apply several accelerations to support large-scale generation.

\textbf{Geometry–specific languages.}
A symbolic geometry engine requires explicit formal languages for geometric constructions and predicate descriptions. The \textbf{construction language} defines geometric elements (e.g., points, lines, and circles) with commands such as \texttt{rectangle A B C D} and \texttt{circle O A B C}. The \textbf{predicate language} describes geometric facts over points, such as congruence (\texttt{cong O A O C}) and parallelism (\texttt{para A B C D}).

\textbf{DDAR engine.}
DDAR combines a rule-based deductive database~\cite{chou2000deductive} with an algebraic reasoning module. The DD component applies rules of the form $P_1,\ldots,P_m \rightarrow P_{\mathrm{new}}$ to derive new predicates, repeating rule matching until no new predicates can be inferred. The AR module maintains quantitative constraints (e.g., lengths and ratios) as linear equations and uses Gaussian elimination to derive additional equalities. Together, DDAR can compute a complete closure of deducible facts.

\textbf{Engine enhancement.}
We enhance the symbolic engine to increase its reasoning capacity through refined deductive rules and an improved arithmetic equation solver.

\textit{Rule Reduction.} The primary motivation for pruning the baseline rule set is to optimize computational efficiency during rule matching. In automated theorem proving, an excessive number of candidate rules at each step can exponentially inflate the search space. Therefore, we eliminate redundant rules that are compositional (i.e., derivable from other rules) or logically trivial (e.g., describing simple algebraic symmetries already implemented by the AR engine).

\textit{Rule Expansion.} Conversely, the introduction of new rules (marked with * in \Cref{tab:deductive_rules}) is driven by the need to enhance geometric expressivity. The baseline rule set treats some metric facts, such as equal distances, in a purely algebraic way. Our new constructive rules (e.g., Circumcenter Definition) serve as semantic bridges that convert these metric constraints into explicit geometric objects (e.g., circles). We also incorporate hierarchical reasoning shortcuts, such as promoting similarity to congruence when a unit scale factor is detected. These rules bypass redundant searches for elementary congruence conditions, such as SSS (Side-Side-Side) and SAS (Side-Angle-Side). Finally, axiomatic rules like Unique Intersection ensure rigorous handling of point-identity scenarios. \Cref{tab:deductive_rules} shows the full rule set used in our GenesisGeo engine.

\newcommand{\backcong}{\mathrel{\reflectbox{$\cong$}}}

\textbf{Notation.}
We use the following notation in the deductive rules:
\begin{itemize}
    \item $\odot{(ABCD)}$: Points $A,B,C,D$ are cyclic.
    \item $\odot{(O;A)}$: Circle with center $O$ and radius $OA$.
    \item $\overline{ABC}$: True if Points $A,B,C$ are collinear.
    \item $\text{Mid}(M,AB)$: Point $M$ is the midpoint of segment $AB$.
    \item $AB:CD$: The ratio of the length of segment $AB$ to the length of segment $CD$.
    \item $A\Rightarrow B$: Means if A is true then B is true.
    \item $\mathcal{O}(ABC)$: Represents the winding order of points $A,B,C$, used to distinguish between direct and reflective symmetry.
    \item $\triangle ABC\sim \triangle DEF$: Two triangles are similar and have the same orientation.
    \item $\triangle ABC\stackrel{\scriptscriptstyle R}{\sim} \triangle DEF$: Two triangles are similar \textbf{as mirror images} (opposite orientation).
    \item $\triangle ABC\cong \triangle DEF$: Two triangles are congruent and have the same orientation.
    \item $\triangle ABC\stackrel{\scriptscriptstyle R}{\cong} \triangle DEF$: Two triangles are congruent \textbf{as mirror images} (opposite orientation).
    \item $\angle ABC = \angle DEF$: The directed angle from line $AB$ to $BC$ is congruent to the one from line $DE$ to $EF$ modulo $\pi$.
    \item $\land$: Logical AND, true only if all connected conditions are true.
    \item $\neg$: Logical NOT, reverses the truth value of a statement.
\end{itemize}

\newcommand{\Col}[1]{\overline{#1}}
\newcommand{\nCol}[1]{\neg \overline{#1}}
\newcommand{\Cyc}[1]{\odot(#1)}
\newcommand{\Cir}[2]{\odot(#1; #2)}
\newcommand{\Mid}[3]{\mathrm{Mid}(#1, #2#3)}
\newcommand{\Land}{\land}   
\newcommand{\Or}[1]{\mathcal{O}(#1)}
\newcommand{\ratio}{{:}}
\newcommand{\eq}{\mkern-2mu=\mkern-2mu}
\newcommand{\noteq}{\mkern-2mu\neq\mkern-2mu}
\newcommand{\para}{\mkern-3mu\parallel\mkern-3mu}
\newcommand{\simT}{\mkern-1.5mu\sim\mkern-1.5mu}
\newcommand{\simR}{\stackrel{\scriptscriptstyle R}{\sim}\mkern-1.5mu}
\newcommand{\congT}{\mkern-2mu\cong\mkern-2mu}
\newcommand{\congR}{\mkern-2mu\stackrel{\scriptscriptstyle R}{\cong}\mkern-2mu}
\begin{table*}[!ht]
\centering
\caption{The full set of deductive rules used in GenesisGeo. Newly introduced rules are marked with an asterisk (*)}
\label{tab:deductive_rules}
\resizebox{\linewidth}{!}{%
    \footnotesize
    \renewcommand{\arraystretch}{1.4}
    \setlength{\tabcolsep}{4pt}

    \begin{tabular}{l l}
    \toprule
    \textbf{Rule Name} & \textbf{Logical Implication}\\
    \midrule
    
    \multicolumn{2}{l}{\textbf{\textit{I. Fundamental Incidence, Lines \& Parallelism}}} \\
    \midrule
    Line Definition by Parallelism & $\Col{\mathit{ABC}} \Rightarrow \mathit{AB} \para \mathit{BC}$ \\
    Collinearity of Parallel Lines & $\mathit{AB} \para \mathit{AC} \Rightarrow \Col{\mathit{ABC}}$ \\
    * Line Intersection Axiom & $\Col{PAB} \Land \Col{QAB} \Land \Col{PCD} \Land \Col{QCD} \Land \nCol{ABD} \Rightarrow P \eq Q$ \\
    Pappus's Hexagon Theorem & $\Col{\mathit{ABC}} \Land \Col{\mathit{PQR}} \Land \Col{XAQ} \Land \Col{XPB} \Land \Col{YAR} \Land \Col{YPC} \Land \Col{ZBR} \Land \Col{ZCQ} \Rightarrow \Col{XYZ}$ \\
    Intercept Theorem & $\mathit{AB} \para \mathit{CD} \Land \Col{OAC} \Land \Col{OBD} \Rightarrow \mathit{OA} \ratio \mathit{OC} \eq \mathit{OB} \ratio \mathit{OD}$ \\
    Converse of Intercept Theorem & $\mathit{OA} \ratio \mathit{OC} \eq \mathit{OB} \ratio \mathit{OD} \Land \Col{OAC} \Land \Col{OBD} \Land \nCol{\mathit{ABC}}  \Rightarrow \mathit{AB} \para \mathit{CD}$ \\
    Intercept Theorem (Trapezoid) & $\mathit{AB} \para \mathit{CD} \Land \Col{MAD} \Land \Col{NBC} \Land \mathit{MA}\ratio \mathit{MD} \eq \mathit{NB} \ratio \mathit{NC} \Rightarrow \mathit{MN} \para \mathit{AB}$ \\
    Proportional Transversals Theorem & $\mathit{AB} \para \mathit{CD} \Land \Col{MAD} \Land \Col{NBC} \Land \mathit{MN} \para \mathit{AB} \Rightarrow \mathit{MA} \ratio \mathit{MD} \eq \mathit{NB} \ratio \mathit{NC}$ \\
    \midrule
    
    \multicolumn{2}{l}{\textbf{\textit{II. Metric Relations \& Circles}}} \\
    \midrule
    Definition of Midpoint & $\mathit{AM} \eq \mathit{BM} \Land \Col{MAB} \Rightarrow \Mid{M}{A}{B}$ \\
    Property of Midpoint & $\Mid{M}{A}{B} \Rightarrow \Col{MAB}$ \\
    Midpoint Ratio Property & $\Mid{M}{A}{B} \Rightarrow \mathit{AM} \ratio \mathit{AB} \eq 1\ratio2$ \\
    * Segment Ratio Property & $\Col{\mathit{ABC}} \Land \Col{DEF} \Land \mathit{AB}\ratio \mathit{AC}\eq\mathit{DE} \ratio \mathit{DF} \Rightarrow \mathit{AB} \ratio \mathit{BC}\eq\mathit{DE} \ratio \mathit{EF}$ \\
    * Radii of Circle & $A,B,C \in \Cir{O}{A} \Rightarrow \mathit{OA} \eq \mathit{OB} \eq \mathit{OC}$ \\
    * Definition of Circumcenter & $\mathit{OA} \eq \mathit{OB} \eq \mathit{OC} \Rightarrow A,B,C \in \Cir{O}{A}$ \\
    Concyclic Equidistance Property & $A,B,C \in \Cir{O}{A} \Land \Cyc{\mathit{ABCD}} \Rightarrow \mathit{OA} \eq \mathit{OD}$ \\
    Equidistance Criterion & $\mathit{OA}\eq \mathit{OB} \Land \mathit{OC}\eq\mathit{OD} \Land \Cyc{\mathit{ABCD}} \Land \mathit{AB} \nparallel \mathit{CD} \Rightarrow \mathit{OA} \eq \mathit{OC}$ \\
    Inscribed Angle Theorem & $\Cyc{\mathit{ABPQ}} \Rightarrow \angle \mathit{APB} \eq \angle \mathit{AQB}$ \\
    Converse of Inscribed Angle Theorem& $\angle \mathit{APB} \eq \angle \mathit{AQB} \Land \nCol{\mathit{ABP}} \Rightarrow \Cyc{\mathit{ABPQ}}$ \\
    * Definition of Secant Line & $\Col{\mathit{PAB}} \Land \mathit{OA} \eq \mathit{OB} \Rightarrow  \Col{\mathit{PAB}} \text{ is a secant of } \Cir{O}{A}$ \\
    \midrule
    
    \multicolumn{2}{l}{\textbf{\textit{III. Triangle Properties}}} \\
    \midrule
    Angle Bisector Theorem & $\mathit{DB} \ratio \mathit{DC} \eq \mathit{AB} \ratio \mathit{AC} \Land \Col{DBC} \Land \nCol{\mathit{ABC}} \Rightarrow \angle \mathit{BAD} \eq \angle \mathit{DAC}$ \\
    Converse of Angle Bisector Theorem & $\angle \mathit{BAD} \eq \angle \mathit{DAC} \Land \Col{DBC} \Land \nCol{\mathit{ABC}} \Rightarrow \mathit{DB} \ratio \mathit{DC} \eq \mathit{AB} \ratio \mathit{AC}$ \\
    Median to Hypotenuse Theorem & $AB \perp BC \Land \Mid{M}{A}{C} \Rightarrow \mathit{AM} \eq \mathit{BM}$ \\
    Concurrency of Altitudes & $\mathit{AB} \perp \mathit{CD} \Land \mathit{AC} \perp \mathit{BD} \Rightarrow \mathit{AD} \perp \mathit{BC}$ \\
    Concurrency of Angle Bisectors & $\angle \mathit{BAX} \eq \angle \mathit{XAC} \Land \angle \mathit{ABX} \eq \angle \mathit{XBC} \Land \nCol{\mathit{ABC}} \Rightarrow \angle \mathit{BCX} \eq \angle \mathit{XCA}$ \\
    \midrule
    
    \multicolumn{2}{l}{\textbf{\textit{IV. Similarity \& Congruence}}} \\
    \midrule
    Definition of Similarity \hfill \makebox[1.0cm][l]{\textit{Direct}} & $\triangle \mathit{ABC} \simT \triangle \mathit{PQR} \Rightarrow \angle \mathit{ABC} \eq \angle \mathit{PQR} \Land \mathit{AB} \ratio \mathit{BC} \eq \mathit{PQ} \ratio \mathit{QR}$ \\
    \phantom{Definition of Similarity} \hfill \makebox[1.0cm][l]{\textit{Reflective}} & $\triangle \mathit{ABC} \simR \triangle \mathit{PQR} \Rightarrow \angle \mathit{ABC} \eq \angle \mathit{PQR} \Land \mathit{AB} \ratio \mathit{BC} \eq \mathit{PQ} \ratio \mathit{QR}$ \\
    
    * Unit Ratio Similarity \hfill \makebox[1.0cm][l]{\textit{Direct}} & $\triangle \mathit{ABC} \simT \triangle \mathit{PQR} \Land \mathit{AB} \eq \mathit{PQ} \Rightarrow \triangle \mathit{ABC} \congT \triangle \mathit{PQR}$ \\
    \phantom{* Unit Ratio Similarity} \hfill \makebox[1.0cm][l]{\textit{Reflective}} & $\triangle \mathit{ABC} \simR \triangle \mathit{PQR} \Land \mathit{AB} \eq \mathit{PQ} \Rightarrow \triangle \mathit{ABC} \congR \triangle \mathit{PQR}$ \\
    
    * CPCTC \phantom{rity Criterion} \hfill \makebox[1.0cm][l]{\textit{Direct}} & $\triangle \mathit{ABC} \congT \triangle \mathit{PQR} \Rightarrow \mathit{AB} \eq \mathit{PQ} \Land \triangle \mathit{ABC} \simT \triangle \mathit{PQR}$ \\
    \phantom{AS Similarity Criterion} \hfill \makebox[1.0cm][l]{\textit{Reflective}} & $\triangle \mathit{ABC} \congR \triangle \mathit{PQR} \Rightarrow \mathit{AB} \eq \mathit{PQ} \Land \triangle \mathit{ABC} \simR \triangle \mathit{PQR}$ \\
    
    AA Similarity Criterion \hfill \makebox[1.0cm][l]{\textit{Direct}} & $\angle \mathit{ABC} \eq \angle \mathit{PQR} \Land \angle \mathit{ACB} \eq \angle \mathit{PQR} \Land \nCol{\mathit{ABC}} \Land \Or{\mathit{ABC}} \eq \Or{\mathit{PQR}} \Rightarrow \triangle \mathit{ABC} \simT \triangle \mathit{PQR}$ \\
    \phantom{AA Similarity Criterion} \hfill \makebox[1.0cm][l]{\textit{Reflective}} & $\angle \mathit{ABC} \eq \angle \mathit{PQR} \Land \angle \mathit{ACB} \eq \angle \mathit{PQR} \Land \nCol{\mathit{ABC}} \Land \Or{\mathit{ABC}} \noteq \Or{\mathit{PQR}} \Rightarrow \triangle \mathit{ABC} \simR \triangle \mathit{PQR}$ \\
    
    SSS Similarity Criterion \hfill \makebox[1.0cm][l]{\textit{Direct}} & $\mathit{AB} \ratio \mathit{BC} \eq \mathit{PQ} \ratio \mathit{QR} \Land \mathit{AC} \ratio \mathit{BC} \eq \mathit{PR} \ratio \mathit{QR} \Land \nCol{\mathit{ABC}} \Land \mathcal{O}(\mathit{ABC}) \eq \mathcal{O}(\mathit{PQR}) \Rightarrow \triangle \mathit{ABC} \simT \triangle \mathit{PQR}$ \\
    \phantom{SSS Similarity Criterion} \hfill \makebox[1.0cm][l]{\textit{Reflective}} & $\mathit{AB} \ratio \mathit{BC} \eq \mathit{PQ} \ratio \mathit{QR} \Land \mathit{AC} \ratio \mathit{BC} \eq \mathit{PR} \ratio \mathit{QR} \Land \nCol{\mathit{ABC}} \Land \mathcal{O}(\mathit{ABC}) \noteq \mathcal{O}(\mathit{PQR}) \Rightarrow \triangle \mathit{ABC} \simR \triangle \mathit{PQR}$ \\
    
    SAS Similarity Criterion \hfill \makebox[1.0cm][l]{\textit{Direct}} & $\mathit{AB} \ratio \mathit{BC} \eq \mathit{PQ} \ratio \mathit{QR} \Land \angle \mathit{ABC} \eq \angle \mathit{PQR} \Land \nCol{\mathit{ABC}} \Land \mathcal{O}(\mathit{ABC}) \eq \mathcal{O}(\mathit{PQR}) \Rightarrow \triangle \mathit{ABC} \simT \triangle \mathit{PQR}$ \\
    \phantom{SAS Similarity Criterion} \hfill \makebox[1.0cm][l]{\textit{Reflective}} & $\mathit{AB} \ratio \mathit{BC} \eq \mathit{PQ} \ratio \mathit{QR} \Land \angle \mathit{ABC} \eq \angle \mathit{PQR} \Land \nCol{\mathit{ABC}} \Land \mathcal{O}(\mathit{ABC}) \noteq \mathcal{O}(\mathit{PQR}) \Rightarrow \triangle \mathit{ABC} \simR \triangle \mathit{PQR}$ \\
    \bottomrule
    \end{tabular}%
}
\end{table*}

\textit{Arithmetic Rules.} While the deductive engine handles qualitative geometric logic, the Algebraic Reasoning (AR) engine provides a quantitative foundation by converting geometric constraints into canonical arithmetic forms. This section details the mapping logic in \Cref{tab:arithmetic equations} and the computational optimizations employed to maintain efficiency within the joint deduction loop.

To bridge the gap between symbolic predicates and numerical constraints, we define a mapping $\mathcal{M}$ that translates high-level geometric relations into algebraic equations (\Cref{tab:arithmetic equations}).
\begin{itemize}
    \item \textbf{Linear Constraints:} Predicates such as \textit{coll} (collinearity) and \textit{para} (parallelism) are linearized using the concept of \textbf{directed angles}. For example, an angle equality $\angle(l_1, l_2) = \angle(l_3, l_4)$ is transformed into:
    \begin{equation*}
        \mathbf{dir}(l_1) + \mathbf{dir}(l_4) - \mathbf{dir}(l_2) - \mathbf{dir}(l_3) = 0 \pmod{\pi}
    \end{equation*}
    \item \textbf{Metric and Complex Constraints:} Distance-based predicates (\textit{cong}, \textit{eqratio}) and complex relations (\textit{secant}) are mapped to logarithmic equations or polynomial identities (e.g., Power of a Point Theorem).
\end{itemize}

Together, these enhancements keep the engine symbolic and efficient, while expanding its deductive closure to cover exponential constraints.

{
\small
\setlength{\tabcolsep}{3pt}
\begin{longtable}{p{0.24\textwidth} >{\ttfamily\raggedright\arraybackslash\small}p{0.72\textwidth}}

\caption{Catalog of Arithmetic Equations} \label{tab:arithmetic equations} \\
\toprule
\textbf{Predicate} & \textbf{\rmfamily Equation} \\
\midrule
\endfirsthead

\caption[]{Catalog of Arithmetic Equations (continued)} \\ 
\toprule
\textbf{Predicate} & \textbf{\rmfamily Equation} \\
\midrule
\endhead


\bottomrule
\endlastfoot

coll A B C & $\angle(AB) = \angle(AC) = \angle(BC)$ \\
& $AB + BC = AC$ \quad \text{(if B lies on segment AC)} \\[2pt]
aconst A B C D r & $\angle(CD) - \angle(AB) = r$ \\[2pt]
cong A B C D & $AB = CD$ \\[2pt]
& $\log(AB) = \log(CD)$ \\
& $\angle(AB) + \angle(AD) = 2 \cdot \angle(BD)$ \quad \text{(if C equals to A)} \\[2pt]
eqangle A B C D E F G H & $\angle(AB) - \angle(cD) = \angle(EF) - \angle(GH)$ \\[2pt]
eqratio A B C D E F G H & $AB \cdot GH = CD \cdot EF$ \\
& $\log(AB) - \log(CD) = \log(EF) - \log(GH)$ \\[2pt]
para A B C D & $\angle(AB) = \angle(CD)$ \\[2pt]
perp A B C D & $\angle(AB) - \angle(CD) = \pi/2$ \\[2pt]
& $AB^2 + AD^2 = BD^2$ \quad \text{(if C equals to A)} \\
rconst A B C D r & $AB = CD \cdot r$ \\[2pt]
secant O A B P & $PA \cdot PB - PO^2 = OA^2 = OB^2$ \quad \text{(if P lies on segment AB)} \\
& $PO^2 - PA \cdot PB = OA^2 = OB^2$ \quad \text{(if P does not lie on segment AB)} \\[2pt]

\end{longtable}
}

\textbf{Engine acceleration.}
A naive implementation of rule matching suffers from poor scalability due to the combinatorial explosion of rule arguments.
To improve scalability, we exploit two acceleration strategies.

\textit{Numeric pre‑identification.} 
Predicates such as \texttt{eqangle} and \texttt{eqratio} require searching over 8-point tuples. We precompute directions for all candidate segments, then keep only tuples that are numerically equal. This step prunes the candidate set before symbolic matching.

\textit{Partial-matching pruning.}
The DD engine matches a rule’s predicates sequentially. If any intermediate predicate cannot be satisfied, we stop matching the remaining predicates. This early-failure pruning avoids exploring impossible combinations and reduces matching cost.

Finally, our optimized symbolic engine is roughly $20\times$ faster than the AlphaGeometry engine.
This speed-up allows us to perform large-scale symbolic deduction within tight time budgets, supporting both dataset generation and downstream search-based proving.

\begin{figure*}[!t]
\centering
\includegraphics[width=1\linewidth]{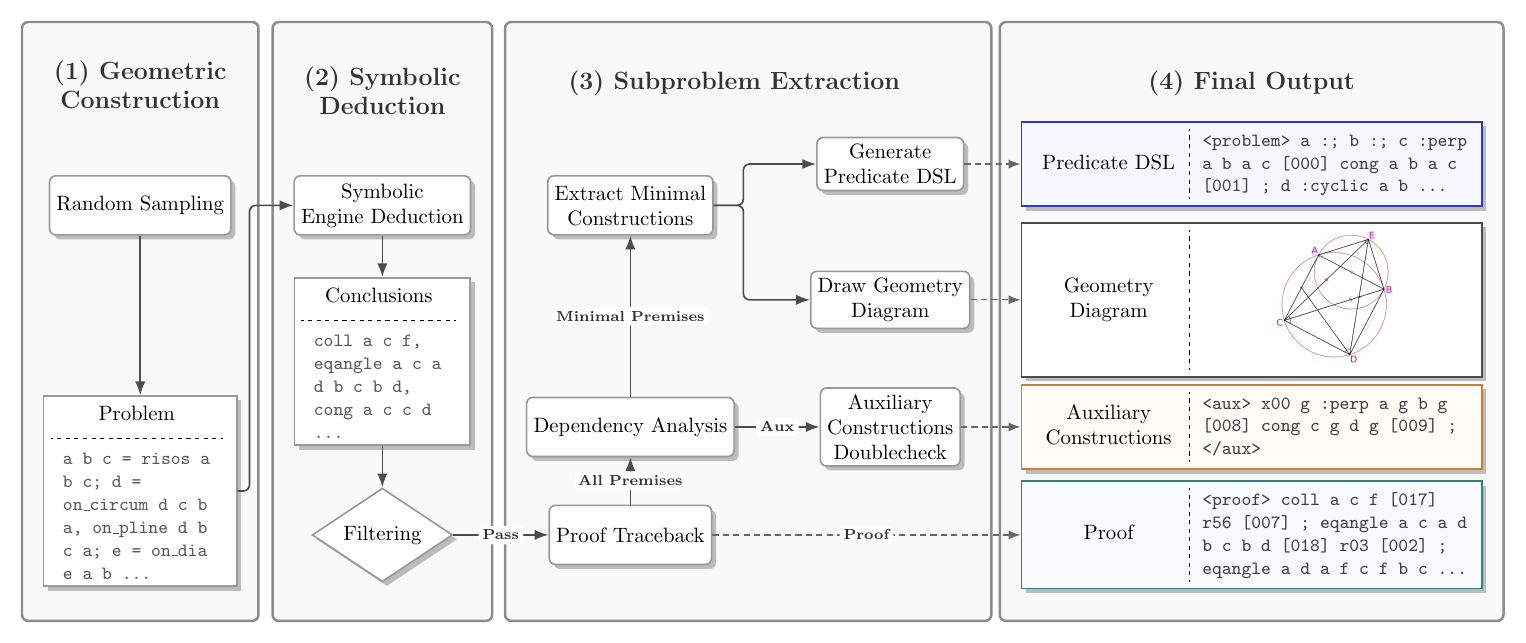} 
\caption{Data synthesis pipeline.
We first sample a construction program in our DSL to instantiate a consistent geometric configuration.
Then, our symbolic engine performs forward deduction from the construction predicates to obtain a closure of conclusions.
From this closure, we select non-trivial targets via predicate-specific filtering and equivalence deduplication.
For each selected goal, we trace back the proof to extract minimal premises and candidate auxiliary constructions, followed by a backward double-check that removes fake auxiliary points.
Finally, we export each problem in a unified multimodal format, including the construction DSL and geometry diagram \mtil{<img>}, the predicate DSL \mtil{<problem>}, a refined auxiliary block \mtil{<aux>}, and a verifiable proof trace \mtil{<proof>}.
}
\label{fig:data_generation_pipeline} 
\end{figure*}

\section{Automated Geometry Data Synthesis}
\label{sec:data_synthesis}

In this section, we build a fully automated synthesis pipeline to create multimodal training data for IMO-level geometry reasoning. \Cref{fig:data_generation_pipeline} summarizes the workflow. Given a sampled construction, the pipeline runs DDAR to compute a closure of deducible statements, selects non-trivial goals through predicate-aware filtering, traces back the dependency graph to extract minimal premises and auxiliary constructions, and renders a diagram consistent with the underlying construction.

\subsection{Geometry Problem Sampling}
We generate candidate problems by sampling action sequences in a constructive DSL adapted from AlphaGeometry~\cite{trinh2024alphageometry}. We organize construction actions into three groups. \texttt{BASIC} initializes an anchor configuration using elementary primitives (e.g., segment, triangle, and rectangle). \texttt{INTERSECT} introduces a new line or circle; pairing two such actions yields a uniquely defined intersection point. \texttt{OTHERS} covers the remaining constructions (e.g., foot, incenter, and excenter).

\textbf{Procedure}. We first sample a \texttt{BASIC} construction to form an initial backbone. Then, we grow the backbone by adding constructions that introduce new lines or circles and their intersections. Concretely, we repeatedly choose either (i) two \texttt{INTERSECT} actions that together determine a new point, or (ii) one \texttt{OTHERS} construction that yields a new object.
Finally, we optionally add intersections among existing geometric objects (e.g., midpoints, feet, reflections, and circle interactions) to increase structural dependency. Note that, before committing each construction, we run dependency checks to ensure prerequisites are available and numerical checks to ensure a consistent coordinate realization exists. These checks prevent invalid actions from propagating and improve generation throughput.

\subsection{Symbolic Deduction and Filtering}
\label{sec:data_forward_filter}

Given a valid problem, we run the symbolic engine to compute a closure of conclusions. This closure contains many predicates that are trivial or equivalent. To ensure that synthesized goals provide meaningful learning signals, we apply a predicate-aware goal filtering procedure.


\textbf{Predicate-specific filtering.}
The symbolic closure contains many candidate goals that are invalid for training.
We filter them with simple, predicate-dependent rules.
We drop trivial statements that hold by definition, such as self-relations (e.g., $AB=AB$ and $AB\parallel AB$).
We also remove goals that can be directly rewritten into a simpler predicate.
For example, $\angle(AB,CD)=0^\circ$ is equivalent to $AB\parallel CD$, and some ratio forms reduce to segment equality.
\Cref{tab:filtering_criteria} lists the exact patterns we exclude for each predicate.

\textbf{Equivalence filtering.}
We further deduplicate high-frequency predicates such as \texttt{eqangle} and \texttt{eqratio}. Two \texttt{eqangle} goals are treated as equivalent when all corresponding line pairs are parallel. Two \texttt{eqratio} goals are treated as equivalent when all corresponding segment pairs are congruent. 

Through two-stage filtering process, we retain meaningful and non-redundant conclusions, ensuring that the resulting problems are both challenging and valuable. 

\begin{table*}[h!]
\centering
\caption{Filtering criteria for geometric conclusions}
\label{tab:filtering_criteria}
\resizebox{\linewidth}{!}{%
\begin{tabular}{lp{8cm}p{5cm}}
\toprule
\textbf{Predicate} & \textbf{Filtered Conclusion Type} & \textbf{Reason for Filtering} \\
\midrule
\texttt{aconst} & $\angle (AB, CD) = 0^\circ$ & Reducible to $AB \parallel CD$ \\
\midrule
\texttt{rconst} & $AB:CD = 1$ & Reducible to $AB = CD$ \\
\midrule
\texttt{cong} & $AB \cong AB$ & Trivial self-congruence \\
\midrule
\texttt{para} & $AB \parallel AB$ & Trivial self-parallelism \\
& $AB \parallel AC$ & Reducible to $A, B, C$ are collinear \\
\midrule
\texttt{eqratio} & $AB/CD = AB/CD$ & Trivial identical ratio \\
& $AB/CD = CD/AB$ & Reducible to $AB = CD$ \\
& $AB/AB = CD/EF$ & Reducible to $CD = EF$ \\
& $AB/CD = EF/GH$ where $AB = CD$ & Reducible to $EF = GH$ \\
& $AB/CD = EF/GH$ where $AB = EF$ & Reducible to $CD = GH$ \\
\midrule
\texttt{eqangle} & $\angle(AB,CD) = \angle(AB,CD)$ & Trivial identical angle \\
& $\angle(AB,CD) = \angle(CD,AB)$ & Reducible to $AB \perp CD$ \\
& $\angle(AB,AB) = \angle(CD,EF)$ & Reducible to $CD \parallel EF$ \\
& Cases involving parallel lines & Reducible to parallel relations \\
& $\angle(AB, CD) = \angle(EF, GH)$ where $AB \perp CD$ & Reducible to $EF \perp GH$ \\
& Cases involving triangle similarity & Reducible to similarity relations \\
\midrule
\texttt{simtri}, \texttt{simtrir} & $\triangle ABC \sim \triangle ABC$ & Trivial self-similarity \\
\midrule
\texttt{contri}, \texttt{contrir} & $\triangle ABC \cong \triangle ABC$ & Trivial self-congruence \\
\bottomrule
\end{tabular}%
}
\end{table*}

\subsection{Subproblem Extraction}
Given a goal, we employ a trace-back algorithm to identify the premises and auxiliary constructions used in the proof. 

\textbf{Premise minimization.} 
Starting from the goal node, we recursively collect only the parent predicates required by each applied rule until reaching construction-derived facts. The resulting leaf set forms the final premise set.

\textbf{Auxiliaries minimization.}
An auxiliary construction is a construction step that is not required to build the geometry problem, but is required by the proof. Concretely, during traceback we identify objects that appear in the proof but cannot be uniquely determined from the extracted premise set. We serialize the construction steps that introduce these objects in the \mtil{<aux>} block, enabling learning settings that predict or verify auxiliary points.

The initial \mtil{<aux>} set may still contain redundant information, because the symbolic engine searches for a shortest deductive path, which may temporarily introduce constructions that are not strictly necessary for proving the target conclusion.
To obtain a minimal auxiliary set, AlphaGeometry~\cite{trinh2024alphageometry} applies a provability check that repeatedly removes points and reruns DDAR, but an exhaustive search over all subsets is exponential and infeasible at scale.
AlphaGeometry2~\cite{chervonyi2025alphageometry2} discards points in a reverse-topological order, which requires only a linear number of checks.

We follow the same pipeline and further improve efficiency by exploiting the fact that many synthesized problems share the same premises and auxiliary candidates but differ only in their target conclusions.
Instead of checking each problem independently, we group problems with identical premises and auxiliary sets and perform provability checks jointly.
In practice, this batching strategy accelerates the auxiliary revalidation process by approximately $5\times$, enabling efficient backward verification at dataset scale.

\subsection{Multimodal Data Format}
Each synthesized example is stored as a single multimodal record that pairs a rendered diagram with formal symbolic annotations and a verifiable derivation.
In practice, one problem consists of four parts:
(1) a \emph{predicate DSL} description that encodes the symbolic premises in a structured \mtil{<problem>} block;
(2) a \mtil{<img>} diagram that visualizes the geometric configuration;
(3) an \mtil{<aux>} block that lists auxiliary constructions required by the proof but not by the original problem; and
(4) a \mtil{<proof>} block that records the full proof trace, including an optional \mtil{<numerical\_check>} section with numerically obtained facts.
Together, these fields provide aligned supervision for diagram understanding, symbolic reasoning, auxiliary construction prediction, and proof generation.

For rendering, we assign Cartesian coordinates to all points during synthesis and draw the diagram from these coordinates. The predicate-based Newclid rendering backend~\cite{sicca2024newclid} fails to visualize certain geometric objects, resulting in diagrams that are not aligned with the symbolic constructions. For instance, the \texttt{triangle} construction defines vertices without predicates to render connecting edges, while a \texttt{circle} is often defined only by two \texttt{cong} predicates (radius equality) without triggering a circle object. To address this, we develop a construction-based rendering system and optimize annotation sizing and color schemes for readability. Finally, we obtain a diagram with precise pixel-level correspondence between symbolic objects and visual elements.

\begin{figure}[!t]
    \centering
    \includegraphics[width=1\linewidth]{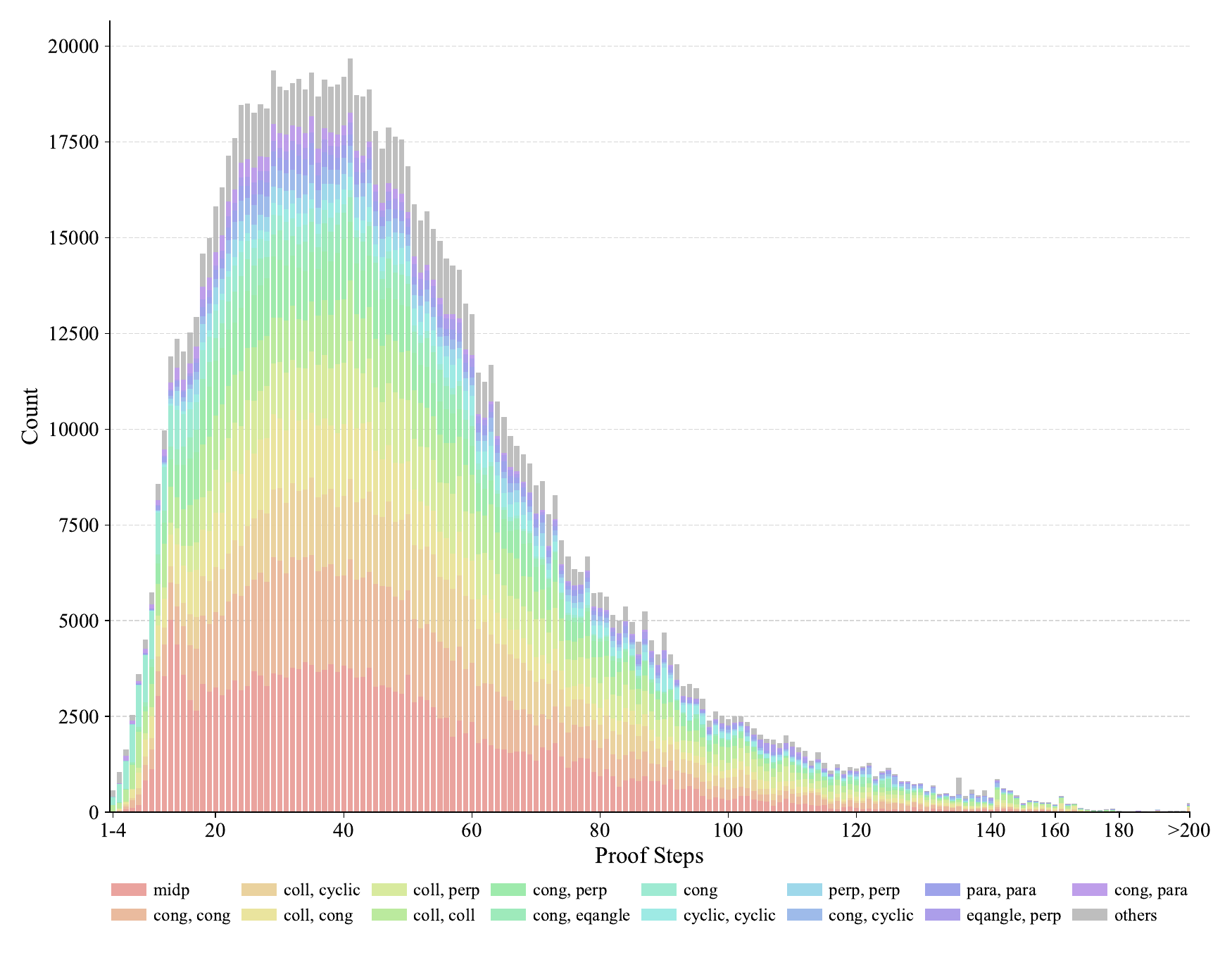}
    \caption{Proof length distribution with auxiliary point types.}
    \label{fig:proofdis}
\end{figure}

\subsection{Dataset Statistics}
\begin{figure*}[!ht]
    \centering
    \includegraphics[width=0.95\linewidth]{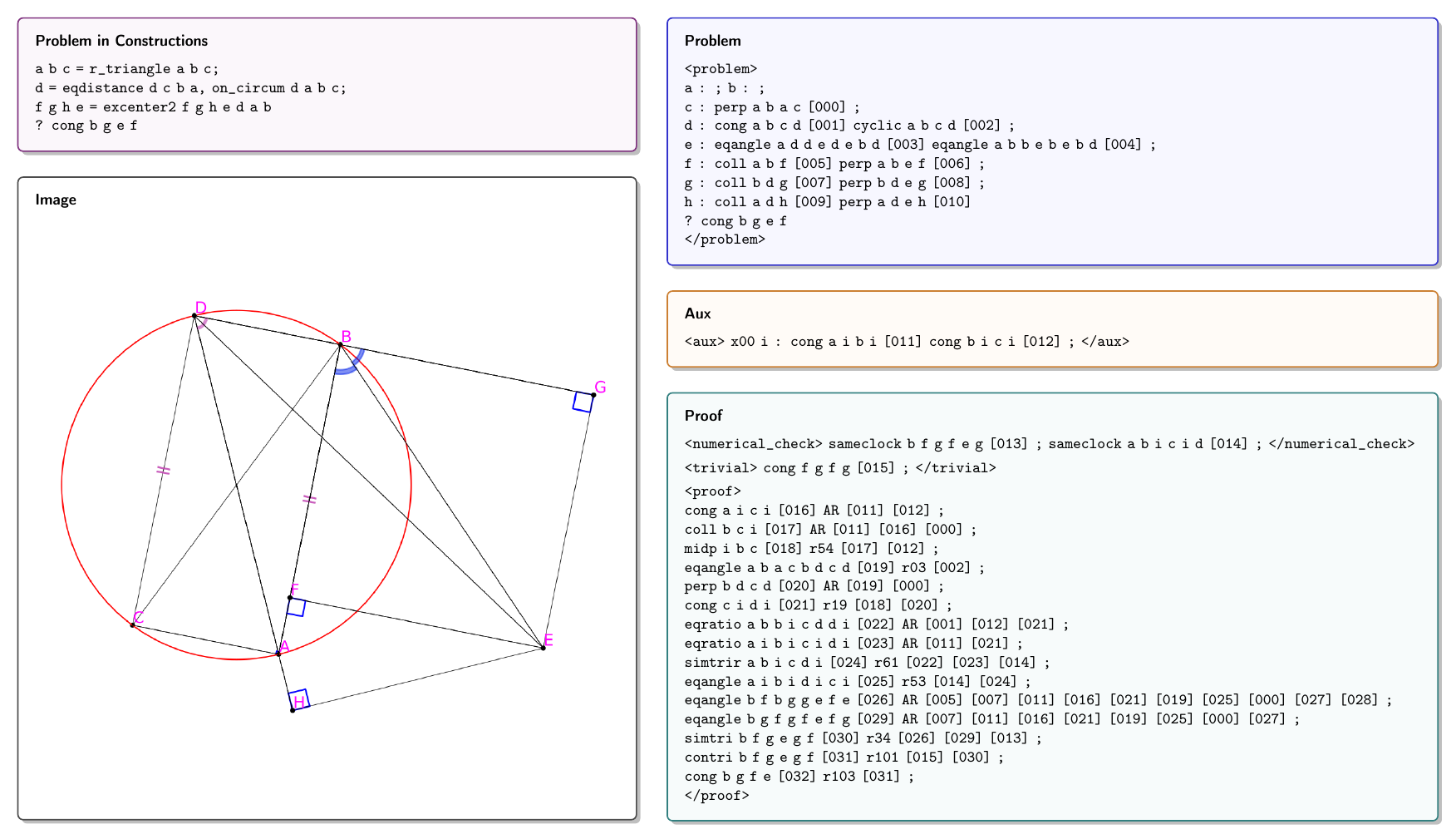}
    \caption{An example of the synthetic data. The \texttt{Problem in Construction} section provides the problem description in the construction DSL, upon which the \texttt{Image} is rendered. The \texttt{Problem} section contains the problem translated into the predicate DSL, detailing the specific geometric properties of each point. The \texttt{<aux>} section specifies the necessary auxiliary constructions. Finally, the \texttt{Proof} block represents the entire deduction process: the \texttt{<numerical\_check>} section contains numerically obtained information and the \texttt{<proof>} session presents the complete sequence of proof steps.}
    \label{fig:example_data}
\end{figure*}

We summarize corpus-level statistics of GenesisGeo-1M in \Cref{fig:proofdis}, which reports the distribution of proof lengths and the composition of the top-15 auxiliary point types.
A problem example is provided in \Cref{fig:example_data}.
Overall, we highlight three properties of the dataset.
\begin{itemize}[itemsep=4pt,topsep=0pt,parsep=0pt,leftmargin=10pt]
    \item \textbf{Large-scale.} The dataset contains 1 million synthesized problems that require auxiliary constructions, generated with 50 CPU threads over 28 hours.
    \item \textbf{Multimodal and verifiable supervision.} Each instance pairs a rendered diagram with formal \emph{construction DSL} and \emph{predicate DSL} descriptions, optional auxiliary constructions, and a complete DDAR proof trace. This format supports joint learning of perception, symbolic reasoning, and proof generation.
    \item \textbf{High-difficulty signals.} Problems are selected from DDAR-derivable but non-trivial goals after predicate-aware filtering and equivalence deduplication. We report difficulty proxies such as the number of premises, proof length and the fraction of instances that require auxiliary constructions, which correlate with longer dependency chains and Olympiad-style reasoning.
\end{itemize}

\section{GenesisGeo: Multimodal Geometry Prover}
We introduce GenesisGeo-2B, a multimodal neuro-symbolic prover that couples a VLM with a symbolic engine.

\subsection{Multi-Task Multimodal Learning}
\begin{figure*}[h]
    \centering
    \includegraphics[width=0.99\linewidth]{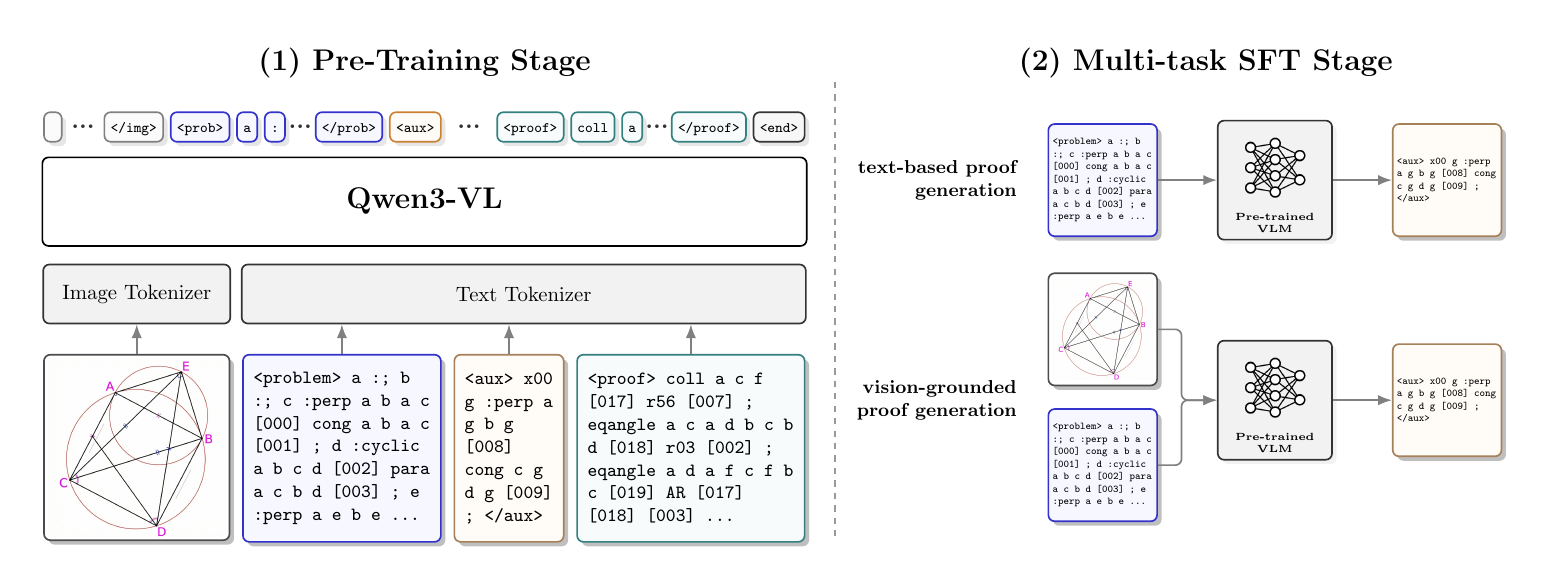} 
    \caption{Multi-stage and multi-task learning framework. We first conduct proof-oriented continued pre-training on complete solution records to align visual diagrams with the predicate language. Then, we  perform supervised multi-task fine-tuning for auxiliary construction, including text-based and vision-grounded settings, enabling robust proposals that integrate symbolic planning with visual grounding.}
    \label{fig:training_process}
\end{figure*}

Our objective is to train the VLM to reliably propose auxiliary constructions that support symbolic deduction across diverse geometric contexts.
We adopt a multi-stage, multi-task learning framework, as illustrated in \Cref{fig:training_process}.

\textbf{Stage I: Alignment-oriented continued pre-training.}
We first perform continued pre-training on complete solution records generated by the symbolic engine.
We frame learning over complete solution records, where the rendered diagram \mtil{<img>}, the formal problem statement \mtil{<problem>}, the auxiliary constructions \mtil{<aux>}, and the corresponding proof trace \mtil{<proof>} jointly serve as both conditioning context and supervision.
During this stage, all components are serialized into a unified sequence and jointly modeled.
This allows the VLM to learn the syntax of predicate-level geometry, the structural patterns of auxiliary constructions, and the global organization of symbolic proofs.
Because all supervision is derived from a single symbolic engine, the model observes consistent reasoning trajectories and avoids learning from heterogeneous or noisy explanations.

\textbf{Stage II: Multi-task supervised fine-tuning.}
We fine-tune the VLM with instruction-tuning tasks that enhance the capabilities needed for auxiliary construction, including text-based and vision-grounded proof generation.
\begin{itemize}[itemsep=2pt,topsep=0pt,parsep=0pt,leftmargin=10pt]
    \item \textbf{Text-based auxiliary construction.} Given predicate-level premises and a goal in \mtil{<problem>}, the model predicts a compact \mtil{<aux>} block. This task enforces symbolic planning without visual cues and trains the model to propose auxiliaries that facilitate symbolic deduction.
    \item \textbf{Vision-grounded auxiliary construction.} Conditioned on both \mtil{<img>} and \mtil{<problem>}, the model proposes a \mtil{<aux>} block that leverages geometric cues in the diagram while respecting the symbolic constraints. This task combines visual grounding with symbolic planning for auxiliary construction.
\end{itemize}

\subsection{Neuro-Symbolic Inference}
GenesisGeo follows an iterative \emph{propose-and-verify} loop for geometry problem solving.
The process begins with the formalized geometry problem. Initially, the symbolic engine attempts to derive the target conclusion directly from the premises.
If the goal is not reachable, the system invokes the visual language model (VLM) to bridge the deductive gap.

\begin{figure*}[!h]
    \centering
    \includegraphics[width=0.95\linewidth]{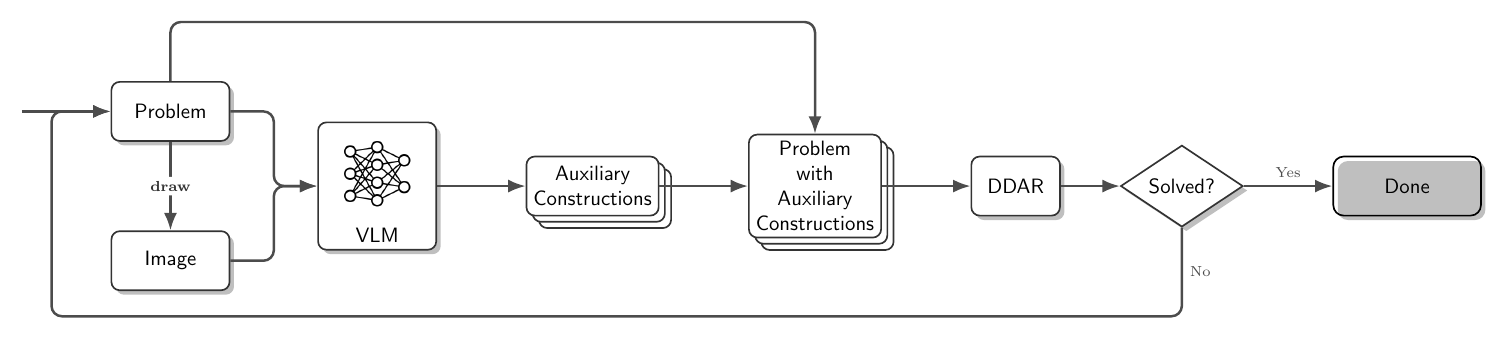}
    \caption{Proving Pipeline of VLM Prover.}
    \label{fig:proving_pipeline}
\end{figure*}

As illustrated in \Cref{fig:proving_pipeline}, our pipeline first renders the problem into a pixel-aligned diagram.
The VLM processes both the symbolic problem statement and the generated image, grounding abstract logical constraints in visual features to propose candidate auxiliary constructions.
These visually informed candidates are then incorporated into the premise set, and the symbolic engine resumes deduction.
This cycle repeats iteratively, allowing the VLM to apply learned perceptual heuristics to guide the symbolic solver through the search space of auxiliary constructions.
Once the goal becomes deducible, the proof is accepted.

In our training corpus, about 90\% of problems require only one auxiliary point, whereas difficult instances demand a deeper search to discover effective constructions. Simply accumulating all proposed auxiliaries in \mtil{<aux>} can deviate from the training distribution. Thus, we adopt an alternative prompting strategy: once an auxiliary point is validated, we fold it into \mtil{<problem>} and re-prompt the VLM to propose the next auxiliary. This keeps \mtil{<aux>} lightweight while improving performance on difficult problems.

To improve efficiency, we employ a parallel beam search strategy in which the language model proposes multiple candidate constructions per iteration, retaining the highest-scoring options for subsequent deduction.
Furthermore, we leverage \texttt{Ray} to establish an asynchronous resource pool: CPU resources occupied by completed problems are dynamically released and reassigned to pending problems, enabling highly parallelized and resource-efficient proof search across large batches.

\section{Experiments}
\label{s:experiments}
\textbf{Training configuration.}
Our GenesisGeo prover is built on the Qwen3-VL-2B architecture~\cite{bai2025qwen3vltechnicalreport} and fine-tuned on \textsc{GenesisGeo-1M} with multi-task training.
Training runs for one epoch with batch size 256 and learning rate $1\times 10^{-4}$ using AdamW~\cite{loshchilov2018decoupled}.
All experiments are conducted on a 64-core CPU machine with eight NVIDIA A100 GPUs.

\textbf{Benchmarks.}
We evaluate our approach on three Euclidean-geometry benchmarks.
\textbf{IMO-30} from AlphaGeometry~\cite{trinh2024alphageometry} contains 30 Olympiad problems.
\textbf{IMO-95} is a subset sourced from the Newclid repository~\cite{sicca2024newclid}, containing formalized IMO problems whose constructions are supported by AlphaGeometry.
\textbf{HAGeo-409} is a larger benchmark of 409 problems covering high-school and Olympiad-level geometry. For a fair comparison across provers, we exclude five problems whose constructions are unsupported by AlphaGeometry, while keeping the original benchmark name.

\textbf{Evaluation configuration.}
We report the number of problems solved under multiple search budgets.
Following HAGeo~\cite{duan2025hagio}, we report three budgets for symbolic engines: \emph{Low (Pass@2048)}, \emph{Medium (Pass@4096)}, and \emph{High (Pass@8192)}.
At each attempt, the solver samples auxiliary points using HAGeo's heuristic, runs the symbolic engine once, and repeats for a fixed number of engine calls.
In addition, we evaluate a tree-search setting for the neuro-symbolic solver, where a large model proposes auxiliary constructions to expand the search tree.
Here, \emph{32$\times$512$\times$4} denotes batch size 32, beam size 512, and proof depth 4, jointly controlling the number of tree-search expansions.
These two budgets are not directly comparable: Pass@K counts engine calls with heuristic sampling, while $32\times512\times 4$ controls the number of model-proposed expansions in a depth-$4$ tree search.
Each solver is allotted a 60-minute time limit per problem.

\begin{table*}[!ht]
\caption{Performance comparison on IMO and HAGeo benchmarks under different search budgets.}
\label{tab:comparison}
\resizebox{\linewidth}{!}{
    \centering
    \setlength{\tabcolsep}{12pt}

    \begin{tabular}{l c c c c c}
    \toprule
    \textbf{Method} & \textbf{Model Size} & \textbf{Search Budget} &
    \textbf{IMO-30} & \textbf{IMO-95} & \textbf{HAGeo-409}\\
    \midrule
    
    \multicolumn{6}{@{}l}{\textit{Symbolic Engines}} \\
    \midrule
    {AG2 (Engine)}                   & - & 0     & 15 & 1    & 87 \\
                                            & - & 2048  & 25 & 41   & 228 \\
                                            & - & 4096  & 25 & 41   & 231 \\
                                            & - & 8192  & 25 & 42   & 231 \\
    \cline{2-6}
    {Newclid 3.0}                    & - & 0     & 16 & 2 & 104 \\
                                            & - & 2048  & 23 & 47 & 241 \\
                                            & - & 4096  & 23 & 50 & 245 \\
                                            & - & 8192  & 23 & 51 & 249 \\\cline{2-6}
    {HAGeo}                          & - & 4096  & 28 & - & - \\
                                            & - & 8192  & - & - & 287 \\
    \cline{2-6}
    {GenesisGeo (Engine)}               & - & 0     & 16 & 2    & 103 \\
                                            & - & 2048  & 26 & 56   & 264 \\
                                            & - & 4096  & 26 & 59   & 268 \\ 
                                            \rowcolor[HTML]{EBF3FF} & - & 8192  & 26 & 59   & 270 \\
    
    \midrule
    
    \multicolumn{6}{@{}l}{\textit{Model-based Provers}} \\
    \midrule
    {AG1} & 151M & 32$\times$512$\times$4 & 25 & 11 & 112 \\
    {AG2} & 3.3B & N/A & 30 & - & - \\
    {TongGeometry} & 1.3B & N/A & 30 & - & - \\
    {SeedProver} & N/A & N/A & 30 & - & - \\
    \rowcolor[HTML]{EBF3FF}
    {GenesisGeo (Text)} & 2B & 32$\times$512$\times$4 & 28 & 59 & 270 \\
    \rowcolor[HTML]{EBF3FF}
    {GenesisGeo (Vision + Text)} & 2B & 32$\times$512$\times$4 & 29 & 63 & 278 \\
    
    \bottomrule
    \end{tabular}
}
\end{table*}

\subsection{Main Results}
\Cref{tab:comparison} reports the number of solved problems under multiple search budgets.
We analyze the results from three perspectives: (1) symbolic engines, (2) neuro-symbolic provers, and (3) failure modes on hard problems.

\textbf{Symbolic engines.} First, we compare pure symbolic engines without auxiliary sampling. Our engine matches Newclid, and both outperform AlphaGeometry2, indicating that our deductive engine is competitive even without search.
We also reimplement HAGeo-style heuristic sampling, where each sampling contains 6 potential auxiliary points.
As we increase the sampling budget from Low to Medium to High, all engines solve more problems.
With heuristic sampling enabled, our engine consistently solves more problems across Low/Medium/High budgets on IMO-30, IMO-95, and HAGeo-409.
This gain is mainly driven by a stronger deductive and arithmetic rule set (\Cref{tab:deductive_rules}).
For example, on IMO-30, our engine solves 26 problems under the High budget, compared to 25 for AlphaGeometry2 and 23 for Newclid.
On HAGeo-409, our engine solves 270 problems under the High budget, compared to 231 for AlphaGeometry2 and 249 for Newclid. However, we cannot directly compare with HAGeo's engine because its code is not publicly available.

\textbf{Neuro-symbolic provers.}
Our solver achieves 29/30 on IMO-30 (\Cref{tab:ag-sg-split}) and maintains strong performance on larger benchmarks.
We evaluate two variants of our model: \emph{GenesisGeo (Text)}, which is fine-tuned only on text-based proof generation tasks, and \emph{GenesisGeo (Vision + Text)}, which is further fine-tuned on both text-based and diagram-grounded tasks.
The result indicates that diagrams provide useful geometric cues, enabling the model to propose more valid auxiliary constructions and thereby improving tree search efficiency.
AlphaGeometry2 and SeedProver do not release their model checkpoints, which prevents a direct comparison.

\textbf{Failure cases.}
The remaining unsolved problems often require deeper auxiliary-construction chains than our current search depth.
For example, AlphaGeometry2 reports that IMO 2008 Problem~6 and IMO 2021 Problem~3 require more than ten auxiliary points, far beyond our default search depth.
We also observe failures when the required construction is rare in the training data or when a small number of early mistakes leads the search into invalid branches.
During development, we found an alternative solution for IMO 2008 Problem~6 that uses only two auxiliary points. The correctness of the proof is confirmed by an IMO
gold medalist. 
To the best of our knowledge, this solution differs from the official solution, the AlphaGeometry2 solution, and other publicly reported solutions.
We include the full proof in the appendix (\Cref{sec:solutions}).

\begin{table*}[!h]
\centering
\setlength{\belowcaptionskip}{1em}
\caption{Performance comparison of AlphaGeometry 1 (AG1) and GenesisGeo (Genesis) on IMO-30.}
\label{tab:ag-sg-split}

\begin{minipage}[t]{0.3\linewidth}
\centering\small
\begin{tabular}{lcc}
\hline
\bfseries ID & \bfseries AG1 & \bfseries Genesis\\
\hline
2000\,P1 & \checkmark & \checkmark\\
2000\,P6 & \checkmark & \checkmark\\
2002\,P2a & \checkmark & \checkmark\\
2002\,P2b & \checkmark & \checkmark\\
2003\,P4 & \checkmark & \checkmark\\
2004\,P1 & \checkmark & \checkmark\\
2004\,P5 & \checkmark & \checkmark\\
2005\,P5 & \checkmark & \checkmark\\
2007\,P4 & \checkmark & \checkmark\\
2008\,P1a & \checkmark & \checkmark\\
\hline
\end{tabular}
\end{minipage}%
\hfill
\begin{minipage}[t]{0.3\linewidth}
\centering\small
\begin{tabular}{lcc}
\hline
\bfseries ID & \bfseries AG1 & \bfseries Genesis\\
\hline
2008\,P1b & $\times$ & \checkmark\\
2008\,P6 & $\times$ & \checkmark\\
2009\,P2 & \checkmark & \checkmark\\
2010\,P2 & \checkmark & \checkmark\\
2010\,P4 & \checkmark & \checkmark\\
2011\,P6 & $\times$ & \checkmark\\
2012\,P1 & \checkmark & \checkmark\\
2012\,P5 & \checkmark & \checkmark\\
2013\,P4 & \checkmark & \checkmark\\
2014\,P4 & \checkmark & \checkmark\\
\hline
\end{tabular}
\end{minipage}%
\hfill
\begin{minipage}[t]{0.3\linewidth}
\centering\small
\begin{tabular}{lcc}
\hline
\bfseries ID & \bfseries AG1 & \bfseries Genesis\\
\hline
2015\,P3 & \checkmark & \checkmark\\
2015\,P4 & \checkmark & \checkmark\\
2016\,P1 & \checkmark & \checkmark\\
2017\,P4 & \checkmark & \checkmark\\
2018\,P1 & \checkmark & \checkmark\\
2019\,P2 & $\times$ & \checkmark\\
2019\,P6 & \checkmark & \checkmark\\
2020\,P1 & \checkmark & \checkmark\\
2021\,P3 & $\times$ & $\times$\\
2022\,P4 & \checkmark & \checkmark\\
\hline
\end{tabular}
\end{minipage}
\end{table*}

\subsection{Ablation Studies}
To understand the contribution of each component, we conduct a series of ablation studies on IMO‑30 and IMO‑95.  

\textbf{Multi-task learning.}
Our multi-task objective includes diagram understanding, text-based proof generation, and diagram-grounded proof generation. To assess each task’s contribution, we train models using single tasks and two-task combinations. Excluding diagram understanding reduces success rates, while removing text-based tasks leads to less precise auxiliary construction suggestions. The full task setting consistently performs best.

\begin{table}[!h]
\caption{Ablation on multi-task training.}
\centering
\small
\setlength{\tabcolsep}{4pt}
\begin{tabular}{lcccc}
\toprule
\textbf{Task mixture} & \textbf{IMO-30} & \textbf{IMO-95} \\
\midrule
Pre-trained Backbone & 27 & 52 \\
Text-based proof generation & 28 & 59 \\
Diagram-based proof generation & 27 & 57  \\
\rowcolor[HTML]{EBF3FF}
All tasks & 29 & 63 \\
\bottomrule
\end{tabular}
\label{tab:ablate_multitask}
\end{table}

\textbf{Training strategy.}
We investigate whether pre-training on \textsc{GenesisGeo} is necessary. We compare (1) models trained only with pre-training (no SFT), (2) models trained from scratch on supervised fine-tuning data only, and (3) models pre-trained on synthetic proofs and then fine-tuned. The results demonstrate that combining pre-training and SFT yields the highest success rates.

\begin{table}[!h]
\caption{Ablation on training strategy. }
\centering
\small
\setlength{\tabcolsep}{4pt}
\begin{tabular}{lccc}
\toprule
\textbf{Training} & \textbf{IMO-30} & \textbf{IMO-95}  \\
\midrule
Pre-training only  & 27 & 52  \\
SFT only & 27 & 57 \\
\rowcolor[HTML]{EBF3FF}
Pre-training + SFT & 29 & 63  \\
\bottomrule
\end{tabular}
\label{tab:ablate_strategy}
\end{table}


\textbf{Search budget.}
We vary the search budget of the neuro-symbolic solver to quantify the trade-off between exploration and compute. Increasing the budget consistently improves the solve rate because the solver can test more candidate auxiliary constructions. Beyond $32\times 512\times 4$, increasing the search budget yields only marginal gains.

\begin{figure}[!ht]
    \centering
    \includegraphics[width=1\linewidth]{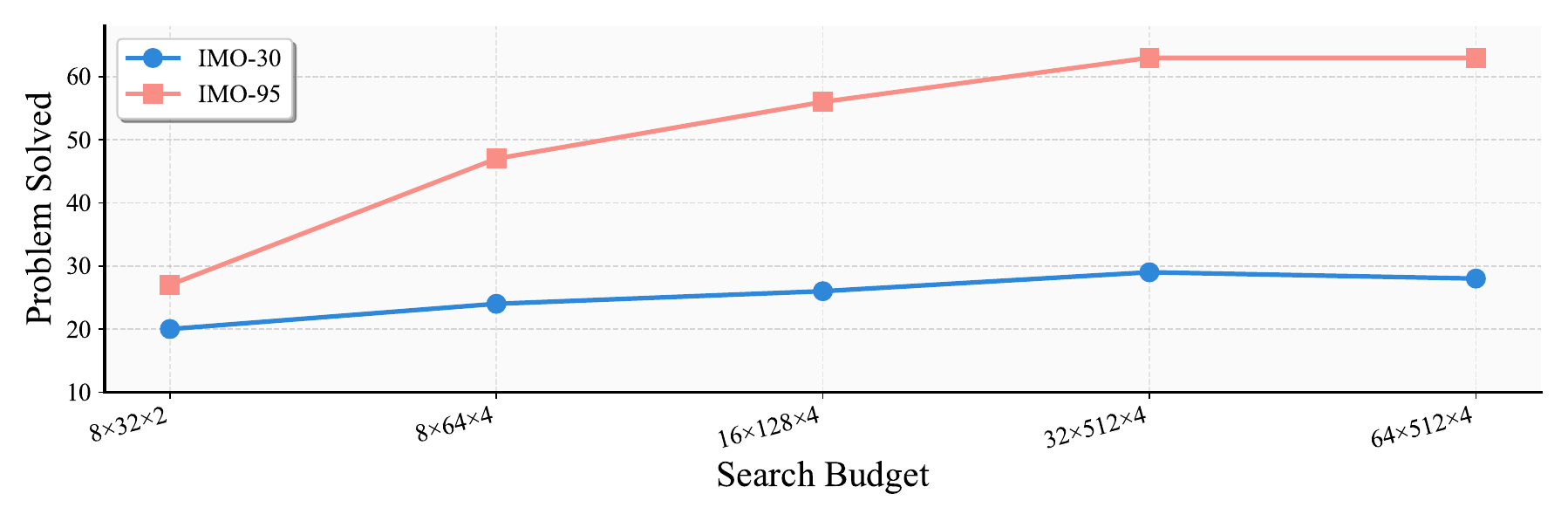}
    \caption{Number of problem solved for different search budgets.}
    \label{fig:searchbudget}
\end{figure}


\section{Conclusion}
We present \textsc{GenesisGeo-1M}, a large-scale synthetic dataset for visual geometric reasoning that pairs a rendered diagram with a verifiable symbolic proof trace. Our data-generation pipeline synthesizes problems that require auxiliary constructions and aligns the symbolic objects used in proofs with the corresponding visual elements, enabling models to exploit diagram evidence while producing checkable proofs.
Building on \textsc{GenesisGeo-1M}, we train multimodal \textsc{GenesisGeo-2B} model with a multi-task learning strategy that combines text-based proof generation and diagram-grounded proof generation. Experiments observe consistent improvements on standard geometry benchmarks. Overall, our results suggest that large-scale multimodal supervision with unified objectives provides a practical path toward connecting diagram understanding with verifiable deduction.
We hope \textsc{GenesisGeo-1M} will support future work on stronger geometry solvers, improved auxiliary-construction search, and more reliable diagram-based reasoning. We will release the dataset, training pipeline, and model checkpoints to the research community.
\clearpage

{
\small
\bibliographystyle{ieeenat_fullname}
\bibliography{main}
}

\newpage
\appendix
\section{Featured GenesisGeo Solutions}
\label{sec:solutions}
During development, we discovered a new solution to IMO 2008 Problem 6, and its correctness was confirmed by an IMO gold medalist. To the best of our knowledge, this solution is different from both the official solution and the one produced by AlphaGeometry2. Notably, AlphaGeometry2\footnote{\url{https://github.com/google-deepmind/alphageometry2}} requires introducing 13 auxiliary points to complete the proof, whereas our solution only requires two auxiliary points.

\begin{problembox}[IMO 2008 P6 (Converted following AlphaGeometry)]
    Let $XYZ$ be a triangle. Define point $O$ as the circumcenter of triangle $XYZ$. Define $\odot{O}$ as the circle centered at $O$ passing through $X$. Let $W$ be any point on $\odot{O}$. Define lines $l_1, l_2, l_3, l_4$ as the tangent lines at $X,Y,Z,W$ on  $\odot{O}$, point $A$ as the intersection of $l_1$ and $l_3$, point $B$ as the intersection of $l_3$ and $l_4$, point $C$ as the intersection of $l_2$ and $l_4$, point $D$ as the intersection of $l_1$ and $l_2$.
    
    Define point $I_1$ as the incenter of triangle $ABC$, point $I_2$ as the incenter of triangle $ACD$, point $F_1$ as the foot of $I_1$ on line $AC$, point $F_2$ as the foot of $I_2$ on line $AC$. Define $\odot{I_1}$ as the circle centered at $I_1$ passing through $F_1$, $\odot{I_2}$ as the circle centered at $I_2$ passing through $F_2$, lines $m_1, m_2$ as the external common tangents of $\odot{I_1}$ and $\odot{I_2}$. Define $P,Q$ as the tangency points of $m_1, m_2$ with $\odot{I_1}$, $S,T$ as the tangency points of $m_1, m_2$ with $\odot{I_2}$, point $K$ as the intersection of lines $m_1$ and $m_2$. \textbf{Prove that:} $OK$ is equal to $OX$ (i.e. $K$ lies on $\odot{O}$).
\end{problembox}

\begin{figure}[htbp]
    \centering
    \includegraphics[width=0.6\linewidth]{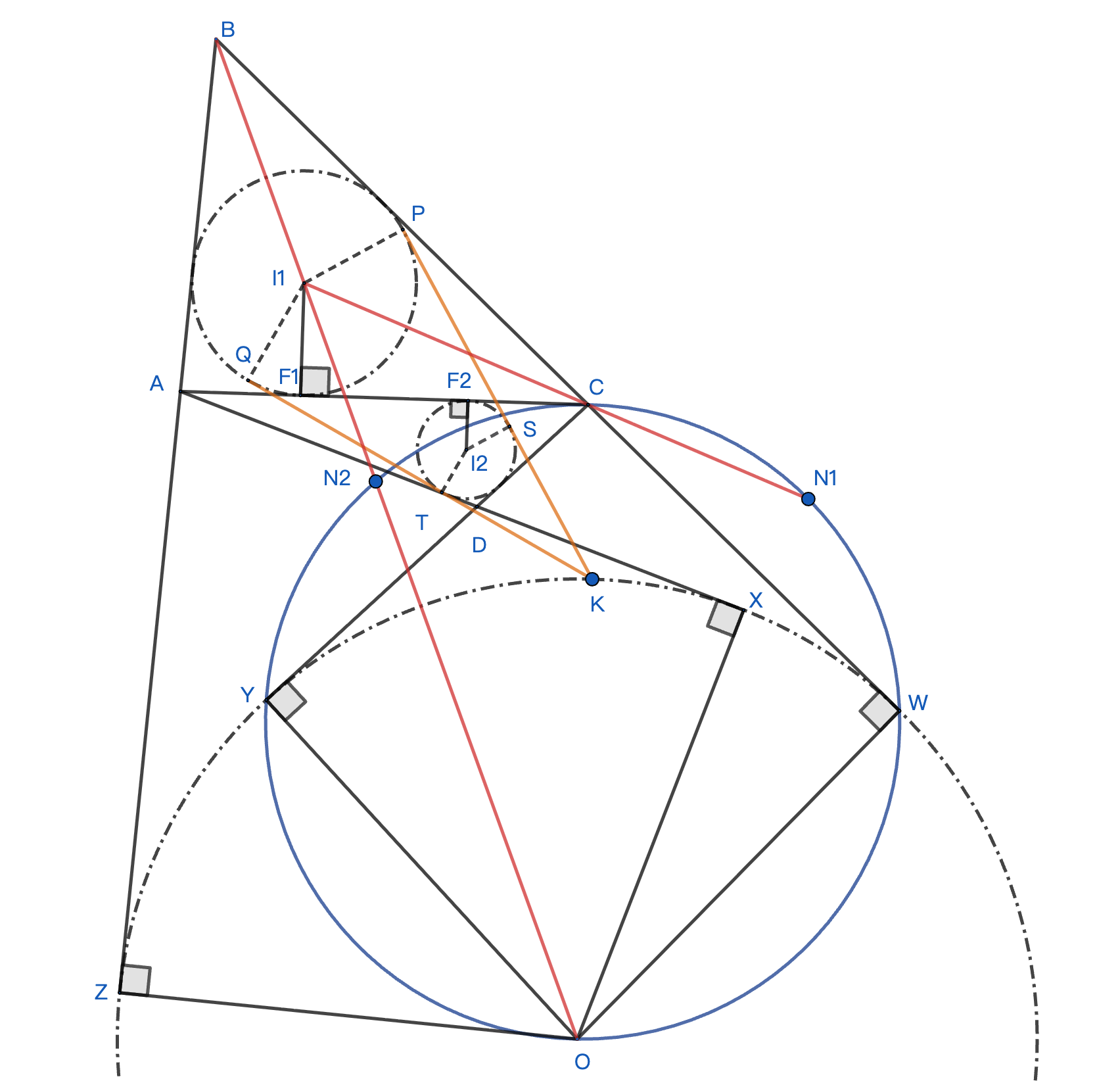}
    
    \caption{Illustration of the problem statement for IMO 2008 P6}
    \label{fig:my_image} 
\end{figure}

\begin{formalbox}[Formal Specification]
\begin{lstlisting}[style=textcode]
Problem Description: 
x y z = triangle x y z; o = circle o x y z; w = on_circle w o x; a = on_tline a z o z, on_tline a x o x; b = on_tline b z o z, on_tline b w o w; c = on_tline c y o y, on_tline c w o w; d = on_tline d x o x, on_tline d y o y; i1 = incenter i1 a b c; i2 = incenter i2 a c d; f1 = foot f1 i1 a c; f2 = foot f2 i2 a c; q t p s = cc_tangent q t p s i1 f1 i2 f2; k = on_line k q t, on_line k p s ? cong o k o x

Our Auxiliary Constructions: 
n1 = on_line n1 c i1, on_circum n1 o w c; n2 = on_line n2 o b, on_circum n2 o w c
\end{lstlisting}
\end{formalbox}

The following trace details the complete derivation process generated by the automated proving engine. It proceeds from the initial premises and auxiliary constructions to the final conclusion through 97 logical steps.

\begin{proofbox}[Full Proof Trace: IMO 2008 P6]
\begin{lstlisting}[style=proofcode]
* From theorem premises:
Points: A, B, C, D, F₁, F₂, I₁, I₂, K, O, P, Q, S, T, W, X, Y, Z
 (Premise)=> OW = OX [0]
 (Premise)=> OX = OY [1]
 (Premise)=> OY = OZ [2]
 (Premise)=> BW ⟂ OW [3]
 (Premise)=> CW ⟂ OW [4]
 (Premise)=> CY ⟂ OY [5]
 (Premise)=> DY ⟂ OY [6]
 (Premise)=> ∠(AC,CI₁) = ∠(CI₁,BC) [7]
 (Premise)=> ∠(AC,AI₂) = ∠(AI₂,AD) [8]
 (Premise)=> ∠(AD,DI₂) = ∠(DI₂,CD) [9]
 (Premise)=> K, P, S are collinear [10]
 (Premise)=> PI₁ = F₁I₁ [11]
 (Premise)=> QI₁ = F₁I₁ [12]
 (Premise)=> SI₂ = F₂I₂ [13]
 (Premise)=> TI₂ = F₂I₂ [14]
 (Premise)=> K, Q, T are collinear [15]
 (Premise)=> PS ⟂ PI₁ [16]
 (Premise)=> QT ⟂ QI₁ [17]
 (Premise)=> PS ⟂ SI₂ [18]
 (Premise)=> QT ⟂ TI₂ [19]
 (Premise)=> AZ ⟂ OZ [20]
 (Premise)=> BZ ⟂ OZ [21]
 (Premise)=> A, C, F₁ are collinear [22]
 (Premise)=> ∠(AB,AI₁) = ∠(AI₁,AC) [23]
 (Premise)=> AC ⟂ F₁I₁ [24]
 (Premise)=> AX ⟂ OX [25]
 (Premise)=> DX ⟂ OX [26]
 (Premise)=> A, C, F₂ are collinear [27]
 (Premise)=> AC ⟂ F₂I₂ [28]
 (Numerical Check)=> A, C, D are not collinear [29]
 (Numerical Check)=> O, W, Y are not collinear [30]
 (Numerical Check)=> P, Q, I₁ are not collinear [31]
 (Numerical Check)=> S, T, I₂ are not collinear [32]
 (Numerical Check)=> O, W, Z are not collinear [33]
 (Numerical Check)=> A, B, C are not collinear [34]
 (Numerical Check)=> AOI₁ are sameclock to CI₂O [35]
 (Numerical Check)=> DOX are sameclock to DYO [36]
 (Numerical Check)=> A, X, Z are not collinear [37]
 (Numerical Check)=> S, I₁, I₂ are not collinear [38]
 (Numerical Check)=> KPQ are sameclock to KST [39]
 (Numerical Check)=> PQI₁ are sameclock to STI₂ [40]
 (Numerical Check)=> K, W, Z are not collinear [41]

* Auxiliary Constructions:
Points: N₁, N₂
 (Premise)=> C, I₁, N₁ are collinear [42]
 (Premise)=> COWN₁ are cyclic [43]
 (Premise)=> B, O, N₂ are collinear [44]
 (Premise)=> COWN₂ are cyclic [45]
 (Numerical Check)=> C, Y, N₁ are not collinear [46]
 (Numerical Check)=> AF₁I₁ are sameclock to CI₁N₂ [47]
 (Numerical Check)=> COI₁ are sameclock to I₁N₁N₂ [48]
 (Numerical Check)=> C, O, N₁ are not collinear [49]
 (Numerical Check)=> CON₁ are sameclock to CF₂I₂ [50]
 (Numerical Check)=> K, N₁, I₂ are not collinear [51]
 (Numerical Check)=> I₁ is to the same side of C->N₁ as I₁ is to K->I₂ [52]
 (Numerical Check)=> C, Y, N₂ are not collinear [53]
 (Numerical Check)=> ACI₁ are sameclock to WN₁N₂ [54]
 (Numerical Check)=> CI₁I₂ are sameclock to KN₁I₁ [55]

Reverse proof steps:
097. KO = OX (by AR with Dist)
╰-- 096. KO = OZ [152] (by circumcenter property r49)
|   |-- 001. O is circumcenter of WYZ [57] (by r72)
|   |   ╰-- 000. OW = OZ [56] (by AR with Dist)
|   |       ╰-- [Premise] OW=OX, OX=OY, OY=OZ
|   ╰-- 095. KWYZ are cyclic [151] (by concyclic check r04)
|       ╰-- 094. ∠(KW,KY) = ∠(WZ,YZ) [150] (by AR with Slope)
|           |-- 093. KN₁ = WN₁ [149] (by AR with Dist)
|           |   ╰-- 092. KN₁:WN₁ = WN₂:ZN₂ [148] (by AR with Distlog "The Grand Ratio")
|           |       |-- 023. KP = KQ [79] (by AR with Slope)
|           |       |-- 024. KS = KT [80] (by AR with Slope)
|           |       |-- 081. WN₂ = ZN₂ [137] (by AR with Slope)
|           |       |   ╰-- 080. Y, Z, N₂ collinear [136] (by AR with Slope)
|           |       |       ╰-- 079. ∠(CO,CY) = ∠(ON₂,YN₂) [135] (by cyclic angles)
|           |       |           ╰-- 078. COYN₂ cyclic [134] (by concyclic check)
|           |       |-- 036. AI₁:CI₁ = F₁I₁:I₁N₂ [92] (by Similar Triangles r53)
|           |       |   ╰-- 035. ▲AF₁I₁ ≅ ▲CN₂I₁ [91] (by Congruence r35)
|           |       |       |-- 034. ∠(AF₁,F₁I₁) = ∠(I₁N₂,CN₂) [90] (by AR with Slope)
|           |       |       |-- 031. ∠(AF₁,AI₁) = ∠(CI₁,CN₂) [87] (by AR with Slope)
|           |       |       ╰-- 047. [Check] Orientation check
|           |       |-- 085. AI₁:CI₁ = WN₁:N₁N₂ [141] (by Similar Triangles r53)
|           |       |   ╰-- 084. ▲ACI₁ ≅ ▲WN₂N₁ [140] (by Congruence r35)
|           |       |       |-- 083. ∠(AI₁,CI₁) = ∠(N₁N₂,WN₁) [139] (by AR with Slope)
|           |       |       |-- 082. ∠(AC,CI₁) = ∠(N₁N₂,WN₂) [138] (by AR with Slope)
|           |       |       ╰-- 054. [Check] Orientation check
|           |       |-- 086. CI₁:CN₁ = I₁I₂:KI₂ [142] (by AR with Dist)
|           |       |   ╰-- 074. CN₁:KI₂ = I₁N₁:KI₁ [130] (by AR with Distlog "The Bridge Ratio")
|           |       |       |-- 065. KP:KS = KI₁:KI₂ [121] (by Thales r07)
|           |       |       |   ╰-- 064. PI₁ ∥ SI₂ [120] (by AR with Slope)
|           |       |       |       ╰-- [16, 18] PS⊥PI₁, PS⊥SI₂
|           |       |       |-- 052. AI₁:CI₂ = OI₁:CO [108] (by Similar Triangles r52)
|           |       |       |   ╰-- 051. ▲AOI₁ ≅ ▲I₂OC [107] (by Congruence r34)
|           |       |       |       |-- 050. ∠(AO,OI₁) = ∠(OI₂,CO) [106] (by AR with Slope)
|           |       |       |       |-- 048. ∠(AO,AI₁) = ∠(OI₂,CI₂) [104] (by AR with Slope)
|           |       |       |       ╰-- 035. [Check] Orientation check
|           |       |       |-- 059. CI₁:OI₁ = I₁N₂:I₁N₁ [115] (by Similar Triangles r53)
|           |       |       |   ╰-- 058. ▲COI₁ ≅ ▲N₂N₁I₁ [114] (by Congruence r35)
|           |       |       |       |-- 057. ∠(CO,OI₁) = ∠(I₁N₁,N₁N₂) [113] (by AR with Slope)
|           |       |       |       |-- 056. ∠(CO,CI₁) = ∠(I₁N₂,N₁N₂) [112] (by AR with Slope)
|           |       |       |       ╰-- 048. [Check] Orientation check
|           |       |       |-- 063. CO:CN₁ = CI₂:F₂I₂ [119] (by Similar Triangles r52)
|           |       |       |   ╰-- 062. ▲CON₁ ≅ ▲I₂CF₂ [118] (by Congruence r34)
|           |       |       |       |-- 061. ∠(CO,CI₂) = ∠(ON₁,CF₂) [117] (by AR with Slope)
|           |       |       |       |-- 060. ∠(CN₁,ON₁) = ∠(F₂I₂,CF₂) [116] (by AR with Slope)
|           |       |       |       ╰-- 050. [Check] Orientation check
|           |       |       |-- 069. KQ:KT = PQ:ST [125] (by Similar Triangles r52)
|           |       |       |   ╰-- 068. ▲KPQ ≅ ▲KST [124] (by Congruence r62)
|           |       |       |       |-- 067. KP:KQ = KS:KT [123] (by AR with Dist)
|           |       |       |       |   ╰-- [79, 80] KP=KQ, KS=KT
|           |       |       |       |-- 066. ∠(KP,KQ) = ∠(KS,KT) [122] (by AR with Slope)
|           |       |       |       ╰-- 039. [Check] Orientation check
|           |       |       ╰-- 073. PQ:PI₁ = ST:TI₂ [129] (by Similar Triangles r53)
|           |       |           ╰-- 072. ▲PQI₁ ≅ ▲TSI₂ [128] (by Congruence r63)
|           |       |               |-- 071. PI₁:QI₁ = TI₂:SI₂ [127] (by AR with Dist)
|           |       |               |   ╰-- [11, 12, 13, 14] Radii equalities
|           |       |               |-- 070. ∠(PI₁,QI₁) = ∠(SI₂,TI₂) [126] (by AR with Slope)
|           |       |               ╰-- 040. [Check] Orientation check
|           |       |-- 090. CI₂:KN₁ = I₁I₂:KI₁ [146] (by Similar Triangles r52)
|           |       |   ╰-- 089. ▲CI₁I₂ ≅ ▲N₁I₁K [145] (by Congruence r62)
|           |       |       |-- 088. CI₁:I₁N₁ = I₁I₂:KI₁ [144] (by AR with Dist)
|           |       |       |   ╰-- 074. CN₁:KI₂ = I₁N₁:KI₁ [130] (See branch 074 above)
|           |       |       |-- 087. ∠(CI₁,I₁N₁) = ∠(I₁I₂,KI₁) [143] (by AR with Slope)
|           |       |       ╰-- 055. [Check] Orientation check
|           |       |-- 091. CO:CI₁ = N₁N₂:I₁N₂ [147] (by Similar Triangles r53)
|           |       |   ╰-- 058. ▲COI₁ ≅ ▲N₂N₁I₁ [114] (See branch 059 above)
|           |       |-- 119. [Using Step 063 result]
|           |       |-- 121. [Using Step 065 result]
|           |       |-- 125. [Using Step 069 result]
|           |       ╰-- 129. [Using Step 073 result]
|           |-- 076. K, Y, N₁ collinear [132] (by AR with Slope)
|           |   |-- 075. CI₂ ∥ KN₁ [131] (by Parallel check r27)
|           |   |   |-- 074. CN₁:KI₂ = I₁N₁:KI₁ [130] (See branch 074 above)
|           |   |   |-- 052. [Check] Orientation check
|           |   |   ╰-- 051. K, N₁, I₂ not collinear [Check]
|           |   |-- 042. C, I₁, N₁ collinear [Premise]
|           |   |-- 058. B, C, W collinear [Step 002]
|           |   |-- 059. C, D, Y collinear [Step 003]
|           |   ╰-- [Premises] BW⊥OW, CW⊥OW, CY⊥OY 
|           |-- 012. COYN₁ cyclic [68] (by concyclic check r04)
|           |   ╰-- 011. ∠(CY,CN₁) = ∠(OY,ON₁) [67] (by AR with Slope)
|           |       |-- 007. ∠(CO,CW) = ∠(ON₁,WN₁) [63] (by cyclic angles)
|           |       |   ╰-- 006. COWN₁ cyclic [Premise 43]
|           |       ╰-- 006. ∠(CO,CN₁) = ∠(OW,WN₁) [62] (by cyclic angles)
|           |           ╰-- 006. COWN₁ cyclic [Premise 43]
|           |-- 010. ∠(CO,CW) = ∠(OY,WY) [66] (by cyclic angles)
|           |   ╰-- 009. COWY cyclic [65] (by concyclic check r04)
|           |       ╰-- 008. ∠(CW,CY) = ∠(OW,OY) [64] (by AR with Slope)
|           |           ╰-- [Premises] CW⊥OW, CY⊥OY
|           ╰-- 005. ∠(AC,CI₂) = ∠(CI₂,CD) [61] (by AR with Slope)
|               ╰-- [Premises] Angle bisectors AI₂, DI₂
╰-- 004. OW = OY [60] (by AR with Dist)
    ╰-- [Premise] OW=OX, OX=OY
\end{lstlisting}
\end{proofbox}
\begin{center}
\end{center}
\newpage
\newcommand{\ang}[2]{\angle(#1, #2)}

\begin{proofbox}[Human-Readable Mathematical Proof of IMO 2008 Problem 6 (GenesisGeo Reconstruction)]

Let $\odot O$ be the circumcircle of $\triangle XYZ$ with center $O$ and radius $R$. Let $W \in \odot O$ be an arbitrary point. Define:
\begin{itemize}
    \item $l_1, l_2, l_3, l_4$: tangent lines to $\odot O$ at $X, Y, Z, W$ respectively
    \item $A = l_1 \cap l_3$, $B = l_3 \cap l_4$, $C = l_2 \cap l_4$, $D = l_1 \cap l_2$
    \item $I_1$: incenter of $\triangle ABC$, $I_2$: incenter of $\triangle ACD$
    \item $F_1$: foot of perpendicular from $I_1$ to $AC$, $F_2$: foot of perpendicular from $I_2$ to $AC$
    \item $\odot I_1$: incircle of $\triangle ABC$ (center $I_1$, radius $r_1 = I_1F_1$),  $\odot I_2$: incircle of $\triangle ACD$ (center $I_2$, radius $r_2 = I_2F_2$)
    \item $m_1, m_2$: external common tangents of $\odot I_1$ and $\odot I_2$
    \item $P, Q$: tangent points of $m_1, m_2$ with $\odot I_1$; $S, T$: tangent points of $m_1, m_2$ with $\odot I_2$
    \item $K = m_1 \cap m_2$
    \item $N_1$: second intersection of ray $CI_1$ with $\odot O$
    \item $N_2$: second intersection of ray $BI_1$ with $\odot O$
\end{itemize}

\vspace{0.3cm}

\textbf{Lemma 1. (The Parallelism Bridge)} \textit{$CI_2 \parallel KN_1$}

\textit{Proof.} We establish the ratio equality:
\begin{equation}
    \frac{CN_1}{KI_2} = \frac{I_1N_1}{KI_1}
    \tag{$\star$}
\end{equation}

Since $K$ is the intersection of external common tangents $m_1, m_2$ of $\odot I_1$ and $\odot I_2$, there exists a homothety $h_K$ centered at $K$ mapping $\odot I_1$ to $\odot I_2$. The tangent points satisfy $I_1P \perp m_1$, $I_2S \perp m_1$, $I_1Q \perp m_2$, $I_2T \perp m_2$.

Since $I_1P \parallel I_2S$ (both perpendicular to $m_1$):
\begin{equation}
    \triangle KPI_1 \sim \triangle KSI_2 \implies \frac{KI_1}{KI_2} = \frac{KP}{KS} = \frac{r_1}{r_2}
    \label{eq:homothety}
\end{equation}

From tangent properties: $PI_1 = QI_1 = r_1$, $SI_2 = TI_2 = r_2$, and $KP = KQ$, $KS = KT$ (tangent segments from external point). The homothety implies:
\begin{equation}
    K, I_1, I_2 \text{ are collinear}
    \label{eq:collinear_K}
\end{equation}

Define $N_2$ as the second intersection of line $BI_1$ with $\odot O$. By the angle bisector property of $I_1$ in $\triangle ABC$:
\begin{equation}
    \angle(AB, BI_1) = \angle(BI_1, BC)
\end{equation}

Since $A, B \in l_3$ (tangent at $Z$) and $B, C \in l_4$ (tangent at $W$), combined with the inscribed angle theorem on $\odot O$, we establish:
\begin{equation}
    B, I_1, N_2 \text{ are collinear} \quad \text{and} \quad O, I_1, N_2 \text{ are collinear}
    \label{eq:collinear_N2}
\end{equation}

From perpendicularity conditions:
\begin{align}
    CW \perp OW, \quad CY \perp OY &\implies C, O, W, Y \text{ are cyclic} \label{eq:cyclic_COWY} \\
    \text{Similarly,} \quad C, O, Y, N_1 &\text{ are cyclic} \label{eq:cyclic_COYN1}
\end{align}

We establish four key ratios through triangle congruences:

\textit{(i)} From $\triangle AF_1I_1 \sim \triangle CN_2I_1$ (proven via angle chasing using \eqref{eq:collinear_N2} and tangent properties):
\begin{equation}
    \frac{AI_1}{CI_1} = \frac{F_1I_1}{I_1N_2} = \frac{r_1}{I_1N_2}
    \label{eq:ratio1}
\end{equation}

\textit{(ii)} From $\triangle AOI_1 \sim \triangle I_2OC$ (proven via angle matching using perpendicularity and \eqref{eq:collinear_N2}):
\begin{equation}
    \frac{AI_1}{CI_2} = \frac{OI_1}{CO}
    \label{eq:ratio2}
\end{equation}

\textit{(iii)} From $\triangle COI_1 \sim \triangle N_2N_1I_1$ (using cyclic quadrilateral $CON_1N_2$ from \eqref{eq:cyclic_COYN1}):
\begin{equation}
    \frac{CI_1}{OI_1} = \frac{I_1N_2}{I_1N_1}
    \label{eq:ratio3}
\end{equation}

\textit{(iv)} From $\triangle CON_1 \sim \triangle I_2CF_2$ (using angle bisector properties and \eqref{eq:cyclic_COYN1}):
\begin{equation}
    \frac{CO}{CN_1} = \frac{CI_2}{r_2}
    \label{eq:ratio4}
\end{equation}

Multiplying \eqref{eq:ratio1} through \eqref{eq:ratio4}:
\begin{equation}
    \frac{AI_1}{CI_1} \cdot \frac{CI_1}{OI_1} \cdot \frac{CO}{CN_1} \cdot \frac{AI_1}{CI_2} = \frac{r_1}{I_1N_2} \cdot \frac{I_1N_2}{I_1N_1} \cdot \frac{CI_2}{r_2} \cdot \frac{OI_1}{CO}
\end{equation}

Simplifying the left side:
\begin{equation}
    \frac{AI_1^2 \cdot CO}{OI_1 \cdot CN_1 \cdot CI_2} = \frac{r_1 \cdot CI_2 \cdot OI_1}{I_1N_1 \cdot r_2 \cdot CO}
\end{equation}

Cross-multiplying and using \eqref{eq:homothety}:
\begin{equation}
    \frac{AI_1^2 \cdot CO^2}{CN_1 \cdot CI_2} = \frac{r_1 \cdot CI_2}{I_1N_1 \cdot r_2}
\end{equation}

From \eqref{eq:ratio2}: $AI_1 \cdot CO = CI_2 \cdot OI_1$. Substituting and simplifying:
\begin{equation}
    \frac{CN_1}{I_1N_1} = \frac{KI_2}{KI_1} \quad \text{(using \eqref{eq:collinear_K} and \eqref{eq:homothety})}
\end{equation}

Since $C, I_1, N_1$ are collinear and $K, I_1, I_2$ are collinear, by Thales' theorem:
\begin{equation}
    \boxed{CI_2 \parallel KN_1}
\end{equation}
\hfill $\square$

\vspace{0.3cm}

\textbf{Lemma 2. (The Isosceles Property)} \textit{$KN_1 = WN_1$}

\textit{Proof.} From the cyclic quadrilaterals $C, O, Y, N_2$ (proven similarly to \eqref{eq:cyclic_COYN1}):
\begin{equation}
    \angle(CO, CY) = \angle(ON_2, YN_2)
\end{equation}

Combined with the tangent properties and the fact that $D, O, I_2$ are collinear (from angle bisector of $\triangle ACD$), the inscribed angle theorem implies:
\begin{equation}
    Y, Z, N_2 \text{ are collinear}
    \label{eq:collinear_YZN2}
\end{equation}

Since $Y, Z, N_2$ are collinear, the chord $YZ$ passes through $N_2$. From the cyclic quadrilateral $B, O, W, Z$ (proven via perpendicularity):
\begin{equation}
    \angle(BW, WZ) = \angle(BO, OZ)
\end{equation}

The configuration exhibits symmetry with respect to the perpendicular bisector of chord $WZ$. Since $N_2$ lies on line $YZ$ and $B, O, N_2$ are collinear, this symmetry yields:
\begin{equation}
    WN_2 = ZN_2
    \label{eq:symmetry}
\end{equation}

From the congruence chain:
\begin{align}
    \triangle ACI_1 &\cong \triangle WN_2N_1 \quad \text{(via angle matching)} \\
    \triangle CI_1I_2 &\cong \triangle N_1I_1K \quad \text{(via Lemma 1 and collinearity)}
\end{align}

We derive:
\begin{equation}
    \frac{AI_1}{CI_1} = \frac{WN_1}{N_1N_2}, \quad \frac{CI_2}{KN_1} = \frac{I_1I_2}{KI_1}
\end{equation}

Combining these with ratios from Lemma 1 and using the distlog (distance-logarithm) algebra:
\begin{equation}
    \frac{KN_1}{WN_1} = \frac{WN_2}{ZN_2}
    \label{eq:grand_ratio}
\end{equation}

Substituting \eqref{eq:symmetry} into \eqref{eq:grand_ratio}:
\begin{equation}
    \frac{KN_1}{WN_1} = 1 \implies \boxed{KN_1 = WN_1}
\end{equation}
\hfill $\square$

\vspace{0.3cm}

\textbf{Lemma 3. (Angle Convergence)} \textit{$\angle(KW, KY) = \angle(ZW, ZY)$, hence $K \in \odot O$}

\textit{Proof.} From Lemma 1: $CI_2 \parallel KN_1$. Since $C, I_1, N_1$ are collinear and $C, O, Y, N_1$ are cyclic:
\begin{equation}
    \angle(CI_2, CY) = \angle(KN_1, N_1Y) \quad \text{(corresponding angles)}
\end{equation}

From the parallelism:
\begin{equation}
    \angle(CI_2, CY) = \angle(KN_1, KY) \quad \text{(alternate angles)}
\end{equation}

Therefore: $\angle(KN_1, N_1Y) = \angle(KN_1, KY)$, which implies:
\begin{equation}
    K, Y, N_1 \text{ are collinear}
    \label{eq:collinear_KYN1}
\end{equation}

From Lemma 2: $KN_1 = WN_1$. Therefore, $\triangle KWN_1$ is isosceles:
\begin{equation}
    \angle(WK, KN_1) = \angle(KW, WN_1) = \alpha
\end{equation}

Since $K, Y, N_1$ are collinear by \eqref{eq:collinear_KYN1}:
\begin{equation}
    \angle(KW, KY) = \angle(KW, KN_1) = \alpha
    \label{eq:angle_KWY}
\end{equation}

Since $W, N_1, Y, Z \in \odot O$ and $Y, N_1$ are collinear with $K$, the inscribed angle theorem gives:
\begin{equation}
    \angle(ZW, ZN_1) = \angle(ZW, ZY) \quad \text{(since $Y, N_1$ collinear with $K$)}
\end{equation}

From the isosceles property of $\triangle KWN_1$ and the fact that $K$ lies on line $YN_1$:
\begin{equation}
    \angle(WN_1, N_1K) = 180^{\circ} - 2\alpha
\end{equation}

By the inscribed angle theorem (angles subtending the same arc $WY$):
\begin{equation}
    \angle(KW, KY) = \angle(ZW, ZY)
    \label{eq:angle_equality}
\end{equation}

By the converse of the inscribed angle theorem: if $\angle(KW, KY) = \angle(ZW, ZY)$, then $K$ lies on the circle through $W, Y, Z$.

Since $W, Y, Z \in \odot O$:
\begin{equation}
    \boxed{K \in \odot O}
\end{equation}
\hfill $\square$

\vspace{0.3cm}

\textbf{Main Theorem.} \textit{$OK = OX$}

\textit{Proof.} From Lemma 3, $K \in \odot O$. Since $\odot O$ is the circumcircle of $\triangle XYZ$ with center $O$ and radius $R$:
\begin{equation}
    OX = OY = OZ = OW = R
\end{equation}

Since $K \in \odot O$:
\begin{equation}
    OK = R = OX
\end{equation}

Therefore:
\begin{equation}
    \boxed{OK = OX}
\end{equation}
\hfill $\blacksquare$

\end{proofbox}

\end{document}